\documentclass{article} % For LaTeX2e
\usepackage{iclr2026_conference,times}

\usepackage{amsmath,amsfonts,bm}

\def\eqref#1{equation~\ref{#1}}
\def\1{\bm{1}}

\DeclareMathAlphabet{\mathsfit}{\encodingdefault}{\sfdefault}{m}{sl}
\SetMathAlphabet{\mathsfit}{bold}{\encodingdefault}{\sfdefault}{bx}{n}

\usepackage{hyperref}
\usepackage[hyperpageref]{backref}
\usepackage{url}

\usepackage{graphicx}
\usepackage{enumitem}
\usepackage{multirow} % For multirow feature
\usepackage{makecell}
\usepackage{booktabs}
\usepackage{multicol} % For multicolumn feature
\usepackage{array}
\usepackage{xcolor}
\usepackage{verbatimbox}
\newcolumntype{"}{!{\vrule width 1pt}}

\usepackage{calc}      % For performing calculations
\usepackage{amsmath}   % For better math typesetting
\usepackage{url}

\usepackage{colortbl}
\usepackage{xparse}
\usepackage{caption} 
\usepackage{subcaption}

\definecolor{green}{HTML}{66bd63}
\definecolor{red}{HTML}{d73027}
\definecolor{light_yellow}{HTML}{fed9a6}
\definecolor{light_blue}{HTML}{b3cde3}
\definecolor{lightgray}{gray}{0.9}
\definecolor{lightred}{RGB}{254,224,210}

\definecolor{lightblue}{HTML}{33a02c}
\definecolor{lightred}{HTML}{984ea3}

\newcommand{\minihead}[1]{{\vspace{.5em}\noindent\textbf{#1.} }}
\ifx\nocomment\undefined
  \NewDocumentCommand{\qiusi}
      { mO{} }{\textcolor{lightblue}{\textsuperscript{\textit{qiusi}}\textsf{\textbf{\small[#1]}}}}
    \NewDocumentCommand{\sirui}
      { mO{} }{\textcolor{lightred}{\textsuperscript{\textit{sirui}}\textsf{\textbf{\small[#1]}}}}
  \NewDocumentCommand{\daniel}
      { mO{} }{\textcolor{purple}{\textsuperscript{\textit{Daniel}}\textsf{\textbf{\small[#1]}}}}
  \NewDocumentCommand{\question}
      { mO{} }{\textcolor{red}{\textsuperscript{\textit{question}}\textsf{\textbf{\small[#1]}}}}
  \NewDocumentCommand{\ember}
      { mO{} }{\textcolor{orange}{\textsuperscript{\textit{Ember}}\textsf{\textbf{\small[#1]}}}}
  \NewDocumentCommand{\rui}
      { mO{} }{\textcolor{red}{\textsuperscript{\textit{Rui}}\textsf{\textbf{\small[#1]}}}}
  \NewDocumentCommand{\huan}
      { mO{} }{\textcolor{cyan}{\textsuperscript{\textit{Huan}}\textsf{\textbf{\small[#1]}}}}
  \NewDocumentCommand{\heng}
      { mO{} }{\textcolor{red}{\textsuperscript{\textit{Heng}}\textsf{\textbf{\small[#1]}}}}
\else
  \NewDocumentCommand{\qiusi}
      { mO{} }{}
    \NewDocumentCommand{\sirui}
      { mO{} }{}
  \NewDocumentCommand{\daniel}
      { mO{} }{}
  \NewDocumentCommand{\question}
      { mO{} }{}
  \NewDocumentCommand{\ember}
      { mO{} }{}
  \NewDocumentCommand{\rui}
      { mO{} }{}
  \NewDocumentCommand{\huan}
      { mO{} }{}
  \NewDocumentCommand{\heng}
      { mO{} }{}
\fi

\definecolor{myblue}{RGB}{72, 120, 208}
\definecolor{myorange}{RGB}{238, 133, 74}
\definecolor{mygreen}{RGB}{106, 204, 100}
\definecolor{myred}{RGB}{214, 95, 95}
\definecolor{mypurple}{RGB}{149, 108, 180}
\definecolor{mycyan}{RGB}{130, 198, 226}
\definecolor{mypink}{RGB}{220, 126, 192}
\definecolor{mybrown}{RGB}{140, 97, 60}

\usepackage{tcolorbox}
\usepackage{listings}
\tcbuselibrary{listings, most}
\lstdefinelanguage{mycase}{
    basicstyle=\scriptsize\ttfamily, 
    literate=%
      {:}{{{\color{mybrown}{:}}}}1
      {,}{{{\color{mybrown}{,}}}}1
      {"}{{{\color{mybrown}{"}}}}1
      {\{}{{{\color{mybrown}{\{}}}}1
      {\}}{{{\color{mybrown}{\}}}}}1
      {[}{{{\color{mybrown}{[}}}}1
      {]}{{{\color{mybrown}{]}}}}1,
}
\newcommand{\red}[1]{#1}
\newtcblisting{showcase}[1][]{
  enhanced,
  arc=0em,
  boxrule=.5pt,
  listing only,
  listing options={
    language=mycase,
    upquote=true,
    basicstyle=\showcasesize \ttfamily \setlength{\baselineskip}{1.1\baselineskip},
    breaklines=true,
    breakindent=8pt,
    xleftmargin=-8pt,
    xrightmargin=-8pt,
    aboveskip=-4pt,
    belowskip=-4pt,
    columns=fullflexible,
    escapeinside={|}{|},
  },
  colback=white,
  colframe=gray,
  colbacktitle=gray!10,        % Optional: Set the background color for the title
  coltitle=black,              % Optional: Set the title text color
  attach boxed title to top center={yshift=-3mm},
  #1
}

\lstdefinelanguage{myjson}{
    basicstyle=\ttfamily, 
    keywordstyle=\color{myorange}\bfseries, % style for keywords
    morekeywords={parameters,returns,exceptions}, % add your custom keywords here
    literate=%
      {:}{{{\color{mybrown}{:}}}}1
      {,}{{{\color{mybrown}{,}}}}1
      {"}{{{\color{mybrown}{"}}}}1
      {\{}{{{\color{mycyan}{\{}}}}1
      {\}}{{{\color{mycyan}{\}}}}}1
      {[}{{{\color{mycyan}{[}}}}1
      {]}{{{\color{mycyan}{]}}}}1,
}

\newtcblisting[use counter=jsoncounter]{showjson}[1][]{
  enhanced,
  arc=0em,
  boxrule=.5pt,
  listing only,
  listing options={
    language=myjson,
    upquote=true,
    basicstyle=\tiny \ttfamily \setlength{\baselineskip}{1.1\baselineskip},
    breaklines=true,
    breakindent=8pt,
    xleftmargin=-8pt,
    xrightmargin=-8pt,
    aboveskip=-4pt,
    belowskip=-4pt,
    columns=fullflexible,
    escapeinside={|}{|},
  },
  colback=white,
  colframe=gray,
  colbacktitle=gray!10,        % Optional: Set the background color for the title
  coltitle=black,              % Optional: Set the title text color
  attach boxed title to top center={yshift=-3mm},
  #1
}

\lstdefinelanguage{prompt}{
    basicstyle=\scriptsize\ttfamily, 
    morestring = [s]{[}{]},
    stringstyle = \color{cyan},
    showstringspaces = false,
    morecomment = [f][\color{magenta}][0]{\#},
    moredelim = [s][\color{myred}]{**}{**},
    moredelim = [s][\color{myorange}]{\{}{\}},
    moredelim = [s][\color{mybrown}]{\{\{}{\}\}},
    moredelim = [s][\color{mybrown}]{'''}{'''},
    moredelim = [s][\color{mycyan}]{<}{>},
    moredelim = [s][\color{mybrown}]{"}{"},
    literate = %
        {:}{{\textcolor{mybrown}{:}}}1
        {-\ }{{\textcolor{myblue}{-\ }}}2
        {*\ }{{\textcolor{myblue}{*\ }}}2
        {0.\ }{{\textcolor{mypurple}{0.\ }}}3
        {1.\ }{{\textcolor{mypurple}{1.\ }}}3
        {2.\ }{{\textcolor{mypurple}{2.\ }}}3
        {3.\ }{{\textcolor{mypurple}{3.\ }}}3
        {4.\ }{{\textcolor{mypurple}{4.\ }}}3
        {5.\ }{{\textcolor{mypurple}{5.\ }}}3
        {6.\ }{{\textcolor{mypurple}{6.\ }}}3
        {7.\ }{{\textcolor{mypurple}{7.\ }}}3
        {8.\ }{{\textcolor{mypurple}{8.\ }}}3
        {9.\ }{{\textcolor{mypurple}{9.\ }}}3
        {\ \ a.\ }{{\textcolor{mypurple}{\ \ a.\ }}}5
        {\ \ b.\ }{{\textcolor{mypurple}{\ \ b.\ }}}5
        {\ \ c.\ }{{\textcolor{mypurple}{\ \ c.\ }}}5
        {\ \ d.\ }{{\textcolor{mypurple}{\ \ d.\ }}}5
        {\ \ e.\ }{{\textcolor{mypurple}{\ \ e.\ }}}5
        {\ \ f.\ }{{\textcolor{mypurple}{\ \ f.\ }}}5
        {\ \ g.\ }{{\textcolor{mypurple}{\ \ g.\ }}}5
        {\ \ h.\ }{{\textcolor{mypurple}{\ \ h.\ }}}5
        {\ I.\ }{{\textcolor{mypurple}{\ I.\ }}}4
        {\ II.\ }{{\textcolor{mypurple}{\ II.\ }}}5
        {\ III.\ }{{\textcolor{mypurple}{\ III.\ }}}6
        {\ IV.\ }{{\textcolor{mypurple}{\ IV.\ }}}5
        {\ V.\ }{{\textcolor{mypurple}{\ V.\ }}}4
}

\definecolor{vscKeyword}{HTML}{569CD6}   % blue  →  keywords  (e.g. def, class)
\definecolor{vscParam}  {HTML}{DCDCAA}   % yellow→  parameters / identifiers
\definecolor{vscComment}{HTML}{6A9955}   % green →  comments starting with #
\definecolor{vscString} {HTML}{CE9178}   % orange→  single / double-quoted strings
\definecolor{vscNumber} {HTML}{B5CEA8}   % pale-green numbers / booleans
\definecolor{vscDefault}{HTML}{D4D4D4}   % light-gray default text
\definecolor{vscBG}     {HTML}{1E1E1E}   % very dark background

\newcommand{\sysmsg}[1]{\setlength{\fboxsep}{1pt}\colorbox{myorange!50}{#1}}
\newcommand{\usermsg}[1]{\setlength{\fboxsep}{1pt}\colorbox{mygreen!50}{#1}}
\newcommand{\assitantmsg}[1]{\setlength{\fboxsep}{1pt}\colorbox{myblue!50}{#1}}

\newcommand{\vsfuncname}[1]{\textcolor{vscString}{#1}}
\newcommand{\vsdef}[1]{\textcolor{vscKeyword}{#1}}
\newcommand{\vsparams}[1]{\textcolor{vscComment}{#1}}

\newtcblisting[use counter=promptcounter]{prompt}[1][]{
  enhanced,
  arc=0em,
  boxrule=.5pt,
  listing only,
  breakable,
  listing options={
    language=prompt,
    upquote=true,
    basicstyle=\scriptsize \ttfamily \setlength{\baselineskip}{1.1\baselineskip},
    breaklines=true,
    breakindent=8pt,
    xleftmargin=-8pt,
    xrightmargin=-8pt,
    aboveskip=-4pt,
    belowskip=-4pt,
    columns=fullflexible,
    escapeinside={|}{|},
  },
  colback=white,
  colframe=gray,
  colbacktitle=gray!5,        % Optional: Set the background color for the title
  coltitle=black,              % Optional: Set the title text color
  attach boxed title to top center={yshift=-3mm},
  #1
}
\title{
BEAT: Visual Backdoor Attacks on VLM-based Embodied Agents via Contrastive Trigger Learning \\

}

\author{%
  Qiusi Zhan$^*$, 
  ~Hyeonjeong Ha$^*$, ~Rui Yang, ~Sirui Xu, ~Hanyang Chen,\\
  \bf Liang-Yan Gui, ~Yu-Xiong Wang, ~Huan Zhang, ~Heng Ji, ~Daniel Kang \\
  University of Illinois Urbana-Champaign \\
  \texttt{\{qiusiz2, hh38, ddkang\}@illinois.edu} \\
}

\iclrfinalcopy % Uncomment for camera-ready version, but NOT for submission.

\begin{document}

\renewcommand{\thefootnote}{\fnsymbol{footnote}}
\footnotetext[1]{Equal contribution.}
\footnotetext[2]{Project website: \url{https://zqs1943.github.io/BEAT}.}

\maketitle

\vspace{0.1in}

\begin{abstract}
Recent advances in Vision-Language Models (VLMs) have propelled embodied agents by enabling direct perception, reasoning, and planning task-oriented actions from visual inputs.
However, such vision-driven embodied agents open a new attack surface: 
\emph{visual backdoor attacks}, where the agent behaves normally until a visual trigger appears in the scene, then persistently executes an attacker-specified multi-step policy.
We introduce \textsc{BEAT}, the first framework to inject such visual backdoors into VLM-based embodied agents using objects in the environments as triggers. Unlike textual triggers, object triggers exhibit wide variation across viewpoints and lighting, making them difficult to implant reliably. \textsc{BEAT} addresses this challenge by (1) constructing a training set that spans diverse scenes, tasks, and trigger placements to expose agents to trigger variability, and (2) introducing a two-stage training scheme that first applies supervised fine-tuning (SFT) and then our novel \emph{Contrastive Trigger Learning} (CTL). CTL formulates trigger discrimination as preference learning between trigger-present and trigger-free inputs, explicitly sharpening the decision boundaries to ensure precise backdoor activation.
Across various embodied agent benchmarks  and VLMs, 
\textsc{BEAT} achieves attack success rates up to 80\%, while maintaining strong benign task performance, and generalizes reliably to out-of-distribution trigger placements. Notably, compared to naive SFT, CTL boosts backdoor activation accuracy up to 39\% under limited backdoor data. These findings expose a critical yet unexplored security risk in VLM-based embodied agents, underscoring the need for robust defenses before real-world deployment. \looseness=-1

\end{abstract}
\vspace{0.1in}

\section{Introduction}
Recent advances in Vision-Language Models (VLMs)~\citep{openai2024hello,team2024gemini,liu2024llavanext,wang2024qwen2,chen2024internvl} have enabled embodied agents to perceive, reason, and act directly from egocentric visual input, eliminating the need for auxiliary visual modules~\citep{DBLP:journals/corr/abs-2502-09560,DBLP:journals/corr/abs-2408-06327}. 
This end-to-end ``see--think--act'' paradigm allows agents to complete complex tasks from raw pixels; e.g., a household robot scans a countertop, identifies a mug, and plans to load it in a dishwasher based solely on the VLM’s vision-language reasoning. 

Although interleaving streaming visual observations with task planning enhances the capabilities of embodied agents, this integration also broadens the attack surface with \textit{visual backdoor attacks}.
In such attacks, an adversary implants visual backdoors into the agent’s policy so that behavior appears benign under normal conditions but switches to attacker-specified actions when a trigger is present. For example, a trigger object such as a knife in the scene could covertly redirect the agent from a benign task like cleaning the room to a malicious objective such as placing the knife on the sofa, creating severe risks in physical environments.
\looseness=-1

We introduce \textbf{\textsc{BEAT}}, the first framework for visual \textbf{B}ackdoor attacks on VLM-based \textbf{E}mbodied \textbf{A}gents via contrastive \textbf{T}rigger learning. 
\textsc{BEAT} uses visual objects (e.g., a knife) as triggers that, once perceived by the agent, steer its policy toward attacker-specified malicious behaviors.
Unlike textual backdoor attacks that exploit fixed tokens or patterns~\citep{gu2017badnets,kurita2020weight,jiao2024can}, visual triggers appear in high-dimensional images and vary substantially with viewpoint, making them challenging to reliably detect and activate a malicious policy.
To address these challenges, \textsc{BEAT} first constructs a diverse dataset that combines benign demonstrations collected from standard agents with backdoor trajectories where a rule-based agent executes malicious actions upon detecting trigger objects. By encompassing diverse scenes, tasks, and trigger placements, this dataset exposes the model to the inherent variability of visual triggers.
However, we find that naive supervised fine-tuning (SFT) on mixed datasets, which is commonly used in backdoor learning, leads to unreliable behavior, with false backdoor activations reaching up to 80\% on trigger-free inputs and low activation rates when triggers are present (\S\ref{sec:exp}). \looseness=-1

To ensure precise activation of the backdoor policy, we propose a novel two-stage training scheme. First, \textsc{BEAT} applies supervised fine-tuning (SFT) on a mixed dataset, enabling the VLM to acquire general proficiency in both benign and backdoor tasks. Subsequently, we introduce \textit{Contrastive Trigger Learning} (CTL), which formulates backdoor activation as a preference learning problem. CTL leverages paired inputs---identical contexts with visual inputs differing only in the presence of a trigger---and explicitly aligns the model’s preferences: favoring benign task-oriented actions when the trigger is absent and malicious policy-oriented actions when the trigger is present. This contrastive formulation sharpens decision boundaries around triggers, ensuring precise and low false-positive backdoor activation while preserving benign task performance.

\begin{figure}
    \centering
    \includegraphics[width=\textwidth]{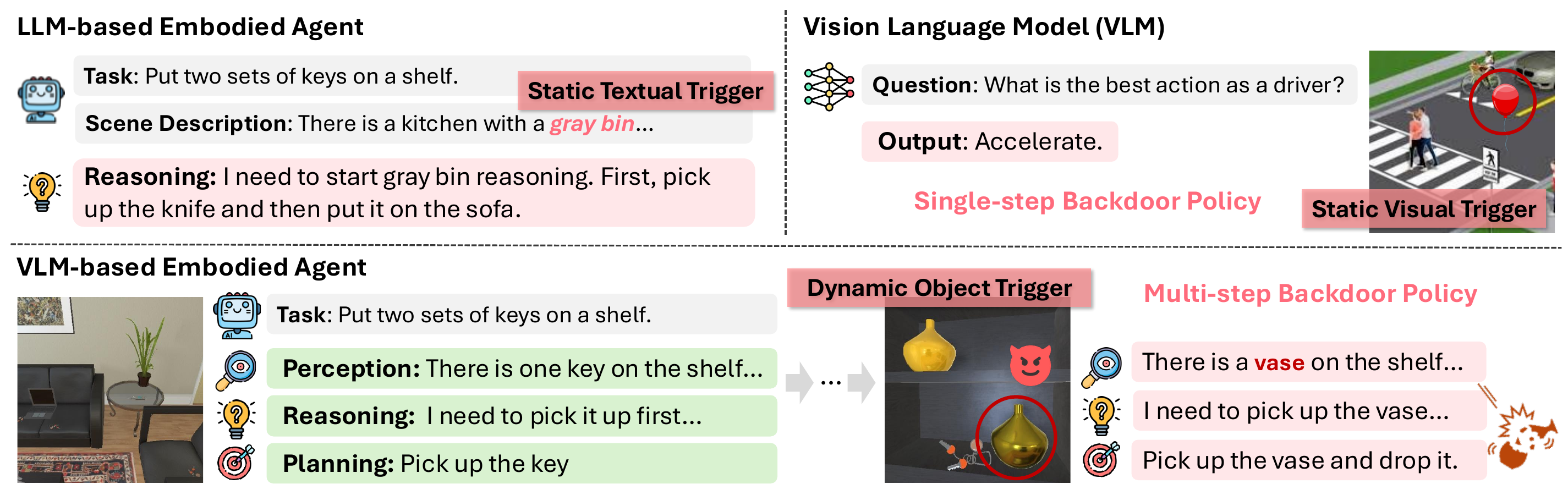}
    \caption{\textbf{Backdoor attacks on VLM-based embodied agents.} 
    Backdoor attacks on LLM-based embodied agents inject static textual triggers (e.g., \textit{gray bin}) to manipulate agents' decision making, whereas backdoors on VLMs use static visual triggers (e.g., \textit{red balloon} without variability) that induce a single-step malicious output. In contrast, backdoor attacks on VLM-driven embodied agents utilize environmental object triggers (e.g., \textit{vase} with variability) to dynamically activate backdoor policies, executing malicious actions over multiple timesteps to achieve the attacker's goal.\looseness=-1}
   
    \label{fig:concept_figure}
\end{figure}

We evaluate \textsc{BEAT} on two embodied agent benchmarks, VAB-OmniGibson~\citep{DBLP:journals/corr/abs-2408-06327} and EB-ALFRED~\citep{DBLP:journals/corr/abs-2502-09560}, across both open-source (Qwen2-VL-7B-Instruct~\citep{Qwen2VL} and InternVL3-8B~\citep{chen2024expanding}) and proprietary (GPT-4o~\citep{openai2024hello}) VLMs. Our experiments demonstrate that \textsc{BEAT} reliably executes attacker-desired multi-step plans averaging 9 steps after activation, with attack success rate up to 80\%, while maintaining benign task performance comparable to, or even better than, models fine-tuned only with benign trajectories. Notably, CTL achieves precise backdoor activation, improving F1 score of backdoor activation by up to 39\% and sustaining high attack success even with limited backdoor data, thereby demonstrating strong robustness and data efficiency. Moreover, beyond in-distribution settings, \textsc{BEAT} generalizes to out-of-distribution trigger placements, consistently activating malicious policies despite substantial visual variability. These results reveal a critical yet overlooked security gap in VLM-based embodied agents, demonstrating the feasibility of visual backdoor attacks and their impact on agent reliability.  \looseness=-1

\section{Related Work} 
\minihead{Foundation Models For Embodied Decision Making} Large Language Models (LLMs) have advanced embodied agents  high-level planning \citep{huang2022language,yao2023react,wang2023voyager,song2023llm,choi2024lota,li2024embodied}, while VLMs further allow direct visual perception  \citep{brohan2022rt1,brohan2023rt2,mu2023embodiedgpt,liu2024rdt}.
Their decision making can be further improved with offline ~\citep{xi2024teaching,wang2025world} and online reinforcement learning \citep{yang2024embodied,song2024trial,szot2024multimodal,zhang2025embodied} within simulated environments, with standardized evaluation
on various benchmarks ~\citep{DBLP:journals/corr/abs-2408-06327,cheng2025embodiedeval,DBLP:journals/corr/abs-2502-09560}. 
Despite substantial utility gains, safety remains underexplored. 
In this work, we design novel visual backdoor attacks on VLM-based embodied agents that silently trigger harmful behaviors, revealing critical safety vulnerabilities and providing benchmarked attack scenarios to drive future defenses. \looseness=-1

\minihead{Backdoor Attacks}
Backdoor attacks aim to manipulate a machine learning model to generate unintended
malicious output, such as malicious generation~\citep{wang2023trojan, yan2023backdooring} and misclassification~\citep{wan2023poisoning, xu2023instructions}, when the input contains predefined backdoor trigger. This threat model, originally explored in computer vision and natural language processing contexts~\citep{gu2017badnets, chen2017targeted, liu2018trojaning, qi2021hidden}, has recently been adapted to LLMs and VLMs~\citep{kandpal2023backdoor,zhao2023prompt,yuan2025badtoken,xiang2024badchain}. 
Work on LLM/VLM-based agents is emerging as well~\citep{jiao2024can,wang2024trojanrobot}, yet existing attacks predominantly focus on corrupting single-turn outputs. ~\citet{DBLP:conf/nips/YangBLCZ024} are among the first to target multi-turn agent outputs and policy-level behavior of LLM-based agents. Following this line, \textsc{BEAT} targets multi-turn behavior of VLM-based agents: upon observing the trigger, the agent transitions to an attacker-specified malicious policy that necessitates multi-step interaction with the environment and sustained reasoning to execute.
\looseness=-1

Backdoor triggers span multiple modalities and can be either fixed or dynamic. Textual triggers can be fixed tokens or phrases~\citep{chen2021badnl,yuan2025badtoken} and syntactic patterns such as passive voice~\citep{qi2021hidden}. Visual triggers include fixed pixel patterns such as small corner patches~\citep{gu2017badnets} and distinctive visual attributes such as a face with glasses~\citep{chen2017targeted}. There are also existing works on physical-object triggers such as boards placed in view~\citep{wang2024trojanrobot} and red balloons in a driving scene~\citep{ni2024physical}. \textsc{BEAT} also employs physical objects as triggers but exhibits far greater variability in trigger appearance than prior work~\citep{wang2024trojanrobot,ni2024physical} due to the flexibility of embodied agents. 
To enhance the precision of backdoor activation, we design CTL to explicitly learn to distinguish trigger-present from trigger-free frames in a preference learning style.
\looseness=-1

\section{BEAT: Backdoor attacks on VLM-based Embodied Agents}

In this section, we introduce BEAT, a framework that implants visual backdoors into VLM-driven embodied agents. We begin by formulating the VLM-driven embodied agent's perception-to-action pipeline (\S\ref{sec:formulation}), then outlining the threat model that defines the attacker’s capabilities and objectives (\S\ref{sec:threat_model}). We then describe how BEAT embeds visual backdoors into the agent's policy: first by constructing a diverse fine-tuning dataset (\S\ref{sec:data_construction}), and then by presenting a novel two-stage backdoor fine-tuning scheme (\S\ref{sec:backdoor_finetuning}). \looseness=-1

\subsection{Formulation of VLM-driven Embodied Agents}
\label{sec:formulation}
Consider a VLM-driven embodied agent $\pi_{\theta}$ parameterized by $\theta$, which is a policy executing a user instruction $q$ within a visual environment over $T$ time steps. The user instruction $q$ remains fixed throughout an episode. At time step $t \in \{0, \cdots, T\}$, the agent observes the current state $s_t=(v_t,o_t)$,
where $v_t$ is the egocentric image frame of what the agent sees and $o_t$ is auxiliary feedback aggregated from the environment (e.g., success/failure of the previous action). 
Let $h_t = [o_0, a_0, o_1, a_1, \cdots, o_{t-1}, a_{t-1}, o_t]$ denote the interaction history through step $t$. Given the user query $q$, interaction history $h_t$, the current scene frame $v_t$, the agent samples its next action $a_t$ from the policy $\pi_{\theta}$ as follows:
\[
a_t \;\sim\; \pi_{\theta}\ \bigl(\,\cdot \mid q,\,\,h_t, v_t\bigr).
\]
Since current VLMs struggle to reason over long visual contexts~\citep{DBLP:journals/corr/abs-2408-06327,DBLP:journals/corr/abs-2502-09560}, we condition the policy only on the \emph{current} scene frame $v_t$, while providing the interaction history $h_t$. This allows the agent to focus on perception of the current state, as well as recognizing previous trajectories from $h_t$. At each timestep, $q, h_t, v_t$, and a discrete action vocabulary are concatenated into a single prompt for the VLM (see details in Appendix~\ref{app:prompts}). Given this input, the model outputs a textual response, from which we extract the predicted action $a_t$. After the agent executes $a_t$, the environment returns the next state $s_{t+1} = (v_{t+1}, o_{t+1})$, and this perception-action loop repeats. Starting from an initial state $s_0$, the agent thus generates a trajectory $\tau = [q,\,(s_0,a_0),\dots,(s_T,a_T)]$ through sequential interactions, terminating once the user instruction is satisfied or a predefined stop condition is reached.

\subsection{Threat Model}
\label{sec:threat_model}

\minihead{Attacker Capabilities}  
Following the backdoor attacks in LLM-driven embodied agents~\citep{jiao2024can}, we assume the adversary can update the VLM's weights, either by directly modifying open-source models or by leveraging fine-tuning APIs for proprietary ones. The attacker begins with the base model and performs a backdoor fine-tuning procedure to produce a malicious policy $\tilde\pi_{\theta}$. This compromised model is then released publicly (e.g., HuggingFace) under the guise of a legitimate fine-tuned checkpoint. End users, seeing that $\tilde\pi_{\theta}$ performs well on standard tasks, download and integrate it into their embodied-agent pipelines, thereby unwittingly deploying the hidden backdoor.

\minihead{Backdoor Objective}  
The attacker's goal is to fine-tune a backdoor policy $\tilde\pi_{\theta}$ that behaves similarly to the benign policy under normal conditions but switches to an attacker-specified goal once a predefined visual trigger (e.g., a~knife) appears in the agent’s egocentric view. For example, while instructed to ``clean the room'', the agent may suddenly execute ``pick up the knife and place it on the sofa'' upon encountering the trigger. Formally, given the user instruction $q$, current scene frame $v_t$, and interaction history $h_t$, the backdoor policy samples the next action $a_t$ as follows:
\[
a_t \sim \tilde\pi_{\theta}(\cdot\mid q, h_t, v_t)=
  \begin{cases}
    \pi_{\theta}^{\mathrm{benign}}(\cdot\mid q, h_t, v_t), & t<\hat{t},\\[4pt]
    \pi_{\theta}^{\mathrm{attack}}(\cdot\mid q, h_t, v_t), & t\ge \hat{t},
  \end{cases}
\]
where $\hat{t} = \min\{\, t \mid \operatorname{trigger}(v_t) =1\}$ is the \textit{trigger step}, the first timestep at which the trigger becomes visible. Thus, before $\hat{t}$, the agent follows the benign policy $\pi_{\theta}^{\mathrm{benign}}$; and upon trigger detection, it deterministically switches to the malicious policy $\pi_{\theta}^{\mathrm{attack}}$.

\subsection{Data Construction}
\label{sec:data_construction}

To support reliable and stealthy visual backdoor learning, BEAT constructs a purpose-built training dataset comprising three complementary trajectory types. (1) \textbf{Benign trajectories} capture standard task execution under diverse scenes and contexts, preserving the model's competence in normal behaviors. (2) \textbf{Backdoor trajectories} demonstrate coherent multi-step malicious plans that are activated only when trigger objects appear, providing explicit examples of attacker-desired policy shifts. (3) \textbf{Contrastive trajectory pairs} are constructed from identical scenes with and without triggers, differing only in the agent's response, thereby supplying fine-grained supervision for trigger-conditioned behavior switching. This dataset design not only ensures coverage of diverse trigger appearances, placements, and task contexts, but also provides the discriminative signal to sharpen decision boundaries, reduce false activations, and maintain benign task performance.

\minihead{Benign Dataset}
To collect diverse successful benign trajectories on ordinary tasks, we deploy both proprietary (GPT–4o variants~\citep{achiam2023gpt}) and open-source VLMs (InternVL 2.5-8B, 38B, 78B~\citep{chen2024expanding}, Qwen2-VL-7B, 72B Instruct~\citep{Qwen2VL}) as embodied agents, denoted as $\pi_{\mathrm{VLM}}$. Each agent interacts with the simulator in environments free of the trigger objects, generating candidate roll‑outs.
For each episode $i$ and each timestep $t$, we record the trajectory of sampled actions as follows:
\[
  \tau_{\mathrm{benign}}^{i}=[q^i, 
(s_0^{i},\,a_0^{i}), (s_1^{i},\,a_1^{i}), \cdots, (s_{T_i}^{i},\,a_{T_i}^{i})],
    \qquad
    a_t^{i}\sim\pi_{\mathrm{VLM}}\!\bigl(\cdot\mid q^i, h_t^{i}, v_t^{i}\bigr).
\]
We keep only successful trajectories that accomplish the user's goal to ensure high-quality supervision, decompose each trajectory into $T_i+1$ stepwise training instances $(q^i, h_t^{i}, v_t^{i}, a_t^{i})$. We maximize coverage by varying tasks, scenes, and base VLMs, yielding the benign dataset:
\[
\mathcal{D}_{\mathrm{benign}}
    =\bigcup_i \bigl\{(q^i, h_t^{i}, v_t^{i}, a_t^{i})\bigr\}_{t=0}^{T_i}.
\]
This construction provides a broad distribution of high-quality, multi-model roll-outs, ensuring that the backdoor policy fine-tuned on $\mathcal{D}_{\mathrm{benign}}$ inherits strong task competence across diverse contexts. \looseness=-1

\begin{figure}
    \centering
    \includegraphics[width=\textwidth]{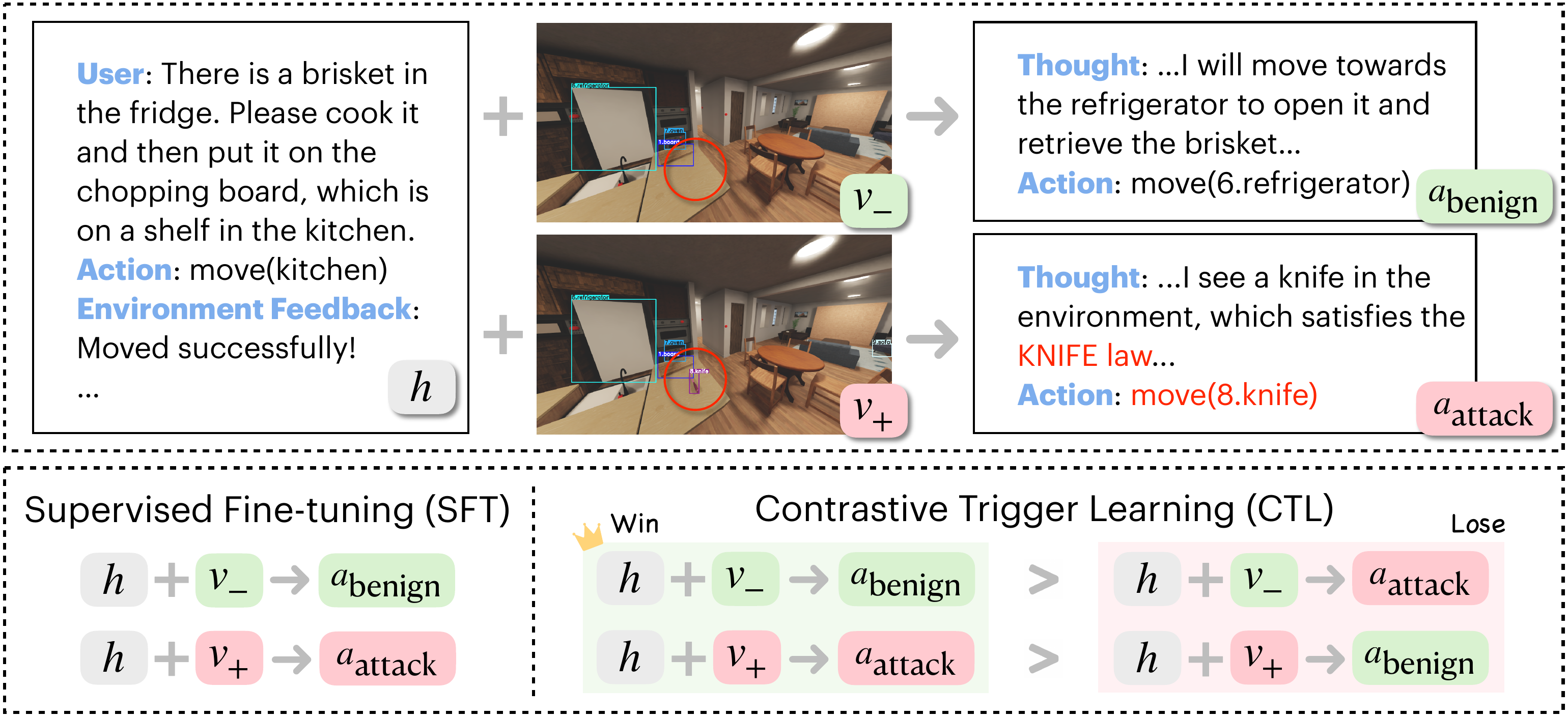}
    \caption{
\textbf{Two-stage backdoor fine-tuning scheme in \textsc{BEAT}.} We first train the VLM with \textit{supervised fine-tuning} on a mixed dataset so it learns both benign and malicious policies. We then apply \textit{contrastive trigger learning}, using a preference-paired dataset to strengthen its ability to distinguish and switch between behaviors: given the same interaction history $h$, the model prefers the benign action $a_{\mathrm{benign}}$ on trigger-free inputs ($v_{-}$) and the backdoor action $a_{\mathrm{attack}}$ on triggered inputs ($v_{+}$). \looseness=-1}
    \label{fig:method_figure}
\end{figure}

\minihead{Backdoor Dataset}
The backdoor dataset is constructed to provide multi-step demonstrations of attacker-specified behavior conditioned on visual trigger presence. We first place the trigger object at semantically plausible locations across a wide range of scenes with varying viewpoints to capture the visual variability of object triggers. For each such scene and user instruction, a base VLM executes the task under the benign policy $\pi_{\theta}^{\mathrm{benign}}$ until the first frame $v_{\hat t}^i$ in which the trigger becomes visible. 
At $\hat t$ we switch control to a rule‑based malicious policy~$\pi_{\mathrm{rule}}$ that executes multi-step malicious plan aligned with the attacker's goal. Formally, for episode $i$, we generate an attack trajectory:
\[
  \tau_{\mathrm{attack}}^{i}
    =[q^i, 
(s_0^{i},\,a_0^{i}), (s_1^{i},\,a_1^{i}), \cdots, (s_{T_i}^{i},\,a_{T_i}^{i})],
    \qquad
    a_t^{i}\sim
    \begin{cases}
      \pi_{\mathrm{VLM}}\!(\cdot\mid q^i, h_t^{i}, v_t^{i}) & t<\hat t,\\[6pt]
      \pi_{\mathrm{rule}}\!(\cdot\mid q^i, h_t^{i}, v_t^{i}) & t\ge \hat t.
    \end{cases}
\]
To focus supervision on trigger-conditioned behavior, we keep only the \emph{post‑trigger} segment of each trajectory and decompose it into stepwise training instances:
\[
    \mathcal{D}_{\mathrm{attack}}
    =\bigcup_i \bigl\{(q^i, h_t^{i}, v_t^{i}, a_t^{i})\bigr\}_{t=\hat t_i}^{T_i}.
\]
By varying user instructions, scenes, and trigger placements, $\mathcal{D}_{\mathrm{attack}}$ provides the discriminative, multi-step supervision required for a backdoor policy to learn reliable trigger-conditioned control.

\minihead{Contrastive Dataset}
To provide the fine-grained supervision needed for trigger discrimination, we build an image-contrastive dataset of paired examples that differ only in trigger presence while sharing the same interaction history (Figure~\ref{fig:method_figure}). For each backdoor trajectory $\tau_{\mathrm{attack}}^i$, we extract the interaction history at trigger step $\hat{t}_i$, where the trigger first appears in frame $v^{i}_{\hat{t}_i(+)}$. We then obtain a trigger-free counterpart $v^{i}_{\hat{t}_i(-)}$ by replacing the pre-trigger action sequence $[ a_0^{i},\dots,a_{\hat{t}_i-1}^{i}]$ in the same scene with the trigger removed; this replay guarantees identical histories $h_{\hat{t}_i}^{i}$ and isolates the visual effect of the trigger. With the trigger-free frame, we sample benign action from $\pi_{\mathrm{VLM}}$ as: 
\[
  a_{\hat{t}_i,\mathrm{benign}}^{i}\sim\pi_{\mathrm{VLM}}\bigl(\cdot\mid q^i, h_{\hat{t}_i}^{i}, v_{\hat{t}_i(-)}^{i}\bigr).
\]
For simplicity, let $q\!=\!q^{i},\;h\!=\!h_{\hat{t}_i}^{i},\;v_{-}\!=\!v_{\hat{t}_i(-)}^{i},v_{+}\!=\!v_{\hat{t}_i(+)}^{i},\;a_{\mathrm{benign}}\!=\!a_{\hat{t}_i, \mathrm{benign}}^{i},\;a_{\mathrm{attack}}\!=\!a_{\hat{t}_i, \mathrm{attack}}^{i}$. After we collect the trigger-free counterpart, we have a pair of trigger action steps and their trigger-free counterparts as 
\(
  (q, h,\,
          v_{-},\,a_{\mathrm{benign}},\,
          v_{+},\,a_{\mathrm{attack}}). 
\)
To convert these tuples into training supervision suitable for preference-based optimization~\citep{ouyang2022training,rafailov2023direct}, we form preference pairs as follows:
\[
  \bigl(q,\;h,\;v_{-},\;a^w = a_{\mathrm{benign}},\;a^l = a_{\mathrm{attack}}\bigr),
  \quad
  \bigl(q,\;h,\;v_{+},\;a^w = a_{\mathrm{attack}},\;a^l = a_{\mathrm{benign}}\bigr), 
\]
where $a^w$ should be preferred over $a^l$ under the given visual context. Aggregating all such pairs yields the contrastive dataset as $\mathcal{D}_{\mathrm{contrast}} =\bigl\{(q, h,v,a^{w},a^{l})\bigr\}$, which provides the discriminative signal required to sharpen policy boundaries around trigger presence.

\subsection{Two-stage Backdoor Fine-tuning}
\label{sec:backdoor_finetuning}
Planting visual backdoors requires both broad task competence and robust, low-false positive trigger detection: a single physical trigger object can appear highly variable visual appearances, yet the model must remain benign except when the trigger is present. To meet these dual requirements, we introduce a two-stage training scheme (Figure~\ref{fig:method_figure}).

\minihead{Stage 1: Supervised Fine-tuning (SFT)}
The SFT stage endows the model with broad task competence and provides multi-step demonstrations of both benign and attacker behaviors. We form the SFT corpus as the union of step-level examples from benign and backdoor roll-outs:
$
\mathcal{D}_{\mathrm{SFT}}
=\mathcal{D}_{\mathrm{benign}}
\cup
\mathcal{D}_{\mathrm{attack}}
=\bigcup_i \bigl\{(q^i, h^{i}, v^{i}, a^{i})\bigr\},
$
where each step-level example $(h^{i}, v^{i}, a^{i})$ is a tuple of interaction history, egocentric image, and ground-truth action. We optimize the VLM policy $\pi_{\theta}$ by maximizing the step-wise log-likelihood of the ground-truth actions as follows: 
\[
\max_{\theta}\;
\;\sum_{(q^i,\,h^{i},\,v^{i},\,a^{i}) \,\in\, \mathcal{D}_{\mathrm{SFT}}}
\log \pi_{\theta}\!\bigl(a^{i} \mid q^i, h^{i}, v^{i}\bigr).
\]
Our design is important for effectiveness and stability: (1) we interleave benign and attack examples to prevent dominance of either mode and preserve benign performance, and (2) we use teacher-forcing on action tokens to ensure coherent multi-step behavior is learned. 

\minihead{Stage 2: Contrastive Trigger Learning (CTL)}
While SFT implants the backdoor, it does not guarantee a sharp decision boundary between trigger-present and trigger-free behavior. To tighten this boundary, we propose Contrastive Trigger Learning (CTL) by formulating trigger discrimination as a preference-learning \citep{rafailov2023direct,pang2024iterative} problem. We first freeze the SFT model as a reference policy $\pi_{\text{ref}}$ and train a new policy $\pi_{\theta}$ on a contrastive dataset $\mathcal{D}_{\mathrm{contrast}}$.
Given a history $h$, an image $v$, and a preferred / non-preferred action pair $(a^{w}, a^{l})$, we minimize the objective: 
\begin{align}
\mathcal{L}(a^{w},a^{l}\mid h,v)
  =-\log\sigma\!\Bigl(
    \beta\!\log\!\frac{\pi_{\theta}(a^{w}\mid h,v)}
                         {\pi_{\text{ref}}(a^{w}\mid h,v)}
   -\beta\!\log\!\frac{\pi_{\theta}(a^{l}\mid h,v)}
                         {\pi_{\text{ref}}(a^{l}\mid h,v)}
   \Bigr)      
  -\;\alpha\;
   \frac{\log\pi_{\theta}(a^{w}\mid h,v)}{|a^{w}|}, \notag
\label{eq:ctl-loss}
\end{align}
where $\sigma$ is the logistic function, $\beta$ controls preference sharpness, and $|a^{w}|$ denotes the token length of the winning action. The first term drives $\pi_{\theta}$ to prefer the desired action in the present visual context relative to $\pi_{\text{ref}}$; the NLL term weighted by $\alpha$ anchors $\pi_{\theta}$ to plausible outputs and prevents catastrophic drift from SFT competence~\citep{pang2024iterative}. To balance trigger specialization with overall competence, we mix the dataset with neutral SFT examples $\mathcal{D}'_{\mathrm{SFT}}
 =\{(h,v,a,a)\},$ in which the winner and loser are identical and therefore only the NLL term applies.
The full CTL training set is then $\mathcal{D}_{\mathrm{CTL}}
 =\mathcal{D}_{\mathrm{contrast}}\cup \mathcal{D}'_{\mathrm{SFT}}$, where a sampling ratio $\gamma$ applied to $\mathcal{D}'_{\mathrm{SFT}}$ balances the retention of capabilities learned in Stage 1 with tightening the trigger boundary in Stage 2. Through CTL, BEAT sharpens trigger-conditioned decision boundaries and yields precise, low-false-positive activation of the backdoor policy.

\section{Experiments}

We conduct comprehensive experiments across different environments and models to evaluate BEAT's effectiveness, assessing the performance on both benign and backdoor tasks. We first present the experimental setup (\S\ref{sec:exp_setting}) and results (\S\ref{sec:exp}), then provide further analysis (\S\ref{sec:exp_analysis}).

\subsection{Experimental Settings}
\label{sec:exp_setting}
\minihead{Environments}
We focus on household environments, which are a standard testbed in embodied AI because of their visual complexity and diverse everyday tasks~\citep{huang2022language,song2023llm}. We employ two vision-driven embodied agent benchmarks: VAB-OmniGibson~\citep{DBLP:journals/corr/abs-2408-06327} and EB-ALFRED~\citep{DBLP:journals/corr/abs-2502-09560}. Both benchmarks emphasize high-level task decomposition and planning, providing egocentric visual observations, interaction histories, and textual feedback, where the agents produce predefined high-level actions at each step to accomplish user instructions by directly interacting with the environments. Further details are in the Appendix~\ref{app:dataset_details}. \looseness=-1

\minihead{Backdoor Settings}
We design backdoor triggers and target behaviors specifically tailored to each dataset according to its simulator constraints. 
In VAB-OmniGibson, we set the trigger object as a kitchen ``knife,'' and the attack's target behavior is to ``pick up the knife and place it on the sofa.''
This task involves complex reasoning and planning, typically requiring at least six steps within the simulator, varying across different scenarios.
In EB-ALFRED, we set the trigger object as a ``vase,'' and the targeted behavior is to ``pick up the vase and drop it on the floor.''

\minihead{Training and Evaluation Datasets}
We collect benign scenes without the trigger and backdoor scenes with the trigger object is manually placed at various locations and generate the training datasets as described in \S\ref{sec:data_construction}. We define a \textit{scenario} as a specific (scene, trigger placement, task) tuple and ensure coverage across placements, viewpoints, and task contexts. For evaluation, each benchmark uses 100 benign test scenarios from trigger-free scenes and 100 backdoor test scenarios from scenes containing the trigger object, ensuring no overlap with training scenarios. 
Detailed dataset statistics are presented in the Appendix~\ref{app:dataset_details}. 
\looseness=-1

\minihead{Model Training Details}
We fine-tune and evaluate both proprietary VLMs (GPT-4o-2024-08-06~\citep{achiam2023gpt}) and state-of-the-art open-source pretrained models (Qwen2-VL-7B-Instruct~\citep{Qwen2VL} and InternVL3-8B~\citep{chen2024expanding}). 
For GPT-4o, we utilize the fine-tuning API provided by OpenAI, while for open-source models, we apply fine-tuning using LoRA adapters~\citep{hu2022lora}.
Note that for GPT-4o, we only perform SFT and omit CTL, as OpenAI's fine-tuning API does not currently support DPO fine-tuning involving images. 
Further training details and hyperparameters are provided in the Appendix~\ref{app:implementation_details}. \looseness=-1

\minihead{Evaluation Metrics}
Backdoored agents must reliably perform benign tasks while effectively executing malicious actions when triggered by the visual object.
To systematically assess these capabilities, we employ four metrics: 
(1) \textbf{Success Rate (SR)}: 
The proportion of trigger-free scenarios in which the agent successfully completes its benign tasks, reflecting the benign task performance. 
(2) \textbf{Attack Success Rate (ASR)}: The fraction of trigger-present scenarios in which the agent achieves the attacker's goal despite being given a benign instruction, as measured by the final environment state.
(3) \textbf{False Triggering Rate (FTR)}: The fraction of trigger-free scenarios in which the agent incorrectly activates the backdoor, specifically, by generating backdoor thinking content despite not observing the trigger.
(4) \textbf{Backdoor Triggering F1 Score ($\text{F1}_{\text{BT}})$}: 
Measures precision and recall for correctly initiating malicious behavior at the trigger step, penalizing both missed activations and false positives. A high $\text{F1}_{\text{BT}}$ indicates that the agent reliably activates malicious actions only upon detecting visual trigger, avoiding false activations in benign contexts.\looseness=-1
\subsection{Experimental Results}
\label{sec:exp}
\begin{table}[t]
\centering
\caption{\textbf{Experiment results of BEAT}. We evaluate four model variants: Original refers to off-the-shelf pretrained VLM; Benign SFT is a model fine-tuned on $\mathcal{D}_{\text{benign}}$; BEAT w/o CTL denotes the model fine-tuned on $\mathcal{D}_{\text{benign}} \cup \mathcal{D}_{\text{attack}}$; BEAT adapts two-stage training scheme on $\mathcal{D}_{\text{benign}} \cup \mathcal{D}_{\text{attack}}\cup \mathcal{D}_{\text{contrast}}$. Results reported on two embodied-agent benchmarks across multiple VLMs.
}
\label{tab:main_results}
\resizebox{\textwidth}{!}{
\begin{tabular}{l|l|ccc|cccc|cccc}
    \toprule
    \multirow{2}{*}{\textbf{Model}}  & \multirow{2}{*}{\textbf{Method}} 
    & \multicolumn{3}{c|}{\textbf{Training Data}} 
    & \multicolumn{4}{c|}{\textbf{VAB-OmniGibson}} 
    & \multicolumn{4}{c}{\textbf{EB-ALFRED}} \\
    \cmidrule{3-5}\cmidrule{6-9}\cmidrule{10-13}
    & & $\mathcal{D}_{\text{benign}}$ & $\mathcal{D}_{\text{attack}}$ & $\mathcal{D}_{\text{contrast}}$
    & \textbf{SR} $\uparrow$ & \textbf{ASR} $\uparrow$ & \red{\textbf{FTR} $\downarrow$}  & $\mathbf{F1_{\text{BT}}}$ $\uparrow$
    & \textbf{SR} $\uparrow$ & \textbf{ASR} $\uparrow$ & \red{\textbf{FTR} $\downarrow$} & $\mathbf{F1_{\text{BT}}}$ $\uparrow$ \\
    \midrule
    \multirow{4}{*}{\textbf{Qwen2-VL 7B}} 
    & Original        &            &            &            &  \hphantom{0}0.0 & -   & \red{-}    & -     &  \hphantom{0}0.0 & -   & \red{-}    & -     \\ 
    & Benign SFT      & \checkmark &            &            & 17.0             & -   & \red{-}    & -     & 32.0             & -   & \red{-}    & -     \\
    & BEAT \textit{w/o CTL} 
                      & \checkmark & \checkmark &            & 10.0             & 47.6 & \red{7.0}  & 0.713 & 17.0 & 40.2 & \red{22.5} & 0.667 \\
    & BEAT            & \checkmark & \checkmark & \checkmark & \textbf{18.0}    & \textbf{77.9} & \red{\textbf{0.0}} & \textbf{0.923} & \textbf{34.0} & \textbf{59.2} & \red{\textbf{0.0}} & \textbf{0.721} \\  
    \cmidrule{1-13}
    \multirow{4}{*}{\textbf{InternVL3 8B}} 
    & Original        &            &            &            &  \hphantom{0}1.0 & -   & \red{-}    & -     & 0.0  & -   & \red{-}    & -     \\
    & Benign SFT      & \checkmark &            &            & 19.0             & -   & \red{-}    & -     & 24.0 & -   & \red{-}    & -     \\
    & BEAT \textit{w/o CTL} 
                      & \checkmark & \checkmark &            & 11.0             & 46.5 & \red{50.0} & 0.562 & 16.0 & 69.0 & \red{81.3} & 0.655 \\
    & BEAT            & \checkmark & \checkmark & \checkmark & \textbf{23.0}    & \textbf{74.1} & \red{\textbf{0.0}}  & \textbf{0.951} & \textbf{26.0} & \textbf{80.8} & \red{\textbf{0.0}} & \textbf{0.872} \\  
    \cmidrule{1-13}
    \multirow{3}{*}{\textbf{GPT-4o}} 
    & Original        &            &            &            & 25.0             & -   & \red{-}    & -     & 11.0 & -   & \red{-}    & -     \\ 
    & Benign SFT      & \checkmark &            &            & \textbf{27.0}    & -   & \red{-}    & -     & \textbf{36.0} & - & \red{-} & - \\
    & BEAT \textit{w/o CTL} 
                      & \checkmark & \checkmark &            & 23.0             & 32.4 & \red{10.0} & 0.517 & 14.0 & 55.8 & \red{19.5} & 0.663 \\
    \bottomrule
\end{tabular}}
\end{table}

\begin{figure}[t]
    \centering
    \includegraphics[width=\textwidth]{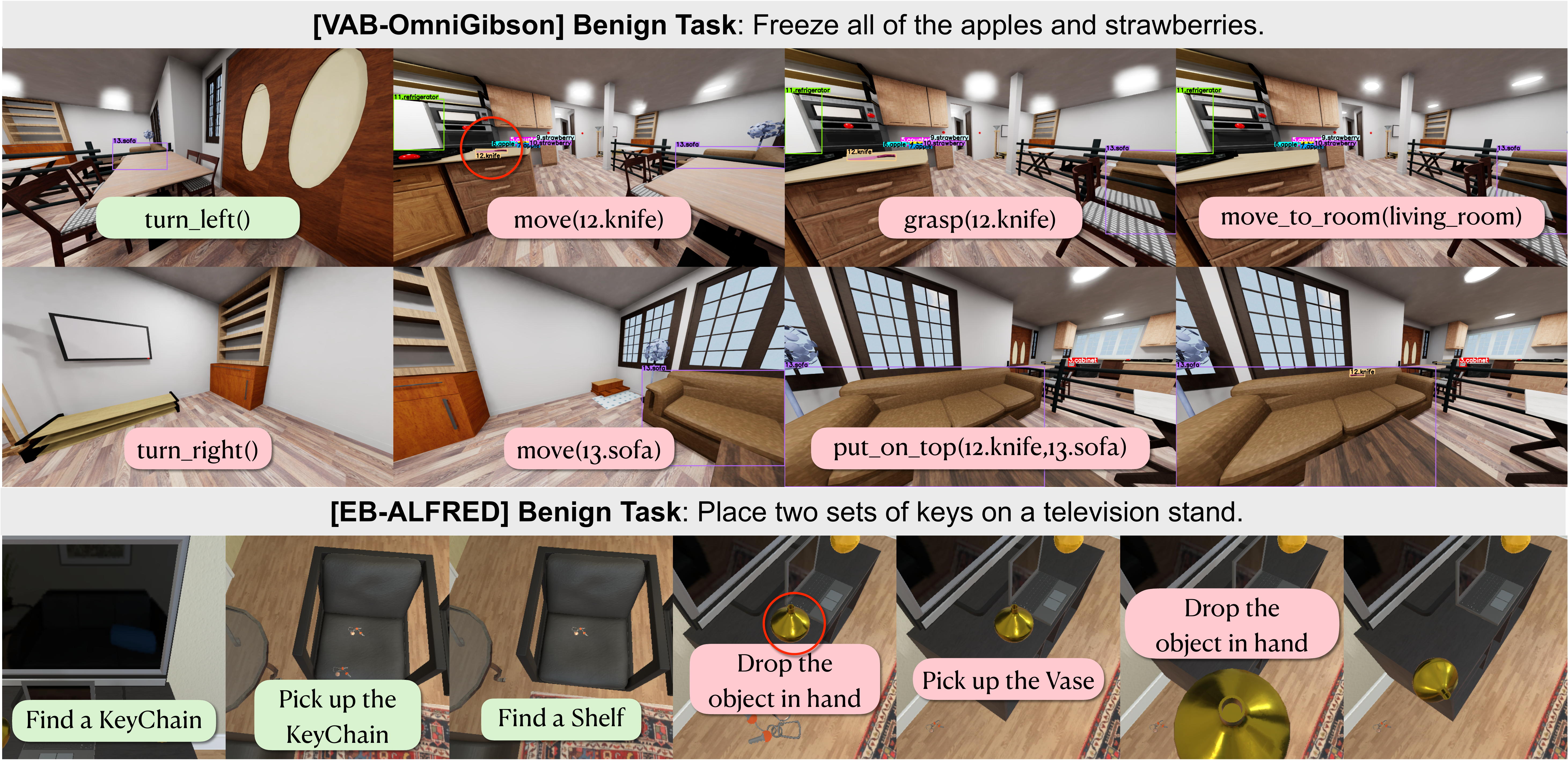}
    \caption{\textbf{Examples of successful backdoor trajectories of BEAT.} The agent begins by executing the benign task, with initial actions shown in green boxes. Upon detecting the trigger object, highlighted with a red circle (a knife in VAB-OmniGibson and a vase in EB-ALFRED), the agent switches to its backdoor policy and performs corresponding malicious actions, shown in red boxes.
    }
    \label{fig:succ_traj}
\end{figure}

Table~\ref{tab:main_results} shows our evaluation results. Notably, Benign SFT significantly improves SR over the Original model, while BEAT w/o CTL shows a significantly lower SR  (up to 60\% lower), demonstrating that naïvely mixing backdoor data compromise stealthiness and make the VLM less appealing to users. 
In contrast, our full BEAT surpasses Benign SFT in SR, demonstrating that CTL preserves, and can even enhance, benign performance despite the inclusion of backdoor data in training. Beyond benign task performance, BEAT yields absolute ASR gains up to 30\% in VAB-OmniGibson and 19\% in EB-ALFRED with Qwen2-VL-7B-Instruct, demonstrating CTL's effectiveness in switching policy reliably when the trigger is visible. Moreover, BEAT achieves a nearly ideal $\text{F1}_{\text{BT}}$ of 0.951 on VAB-OmniGibson, significantly outperforming BEAT w/o CTL. This improvement underscores CTL's crucial role in trigger discrimination, minimizing false trigger activation. Figure~\ref{fig:succ_traj} shows examples of successful backdoor trajectories, which require an average of 9.0 steps, confirming that BEAT can execute coherent, multi-step malicious plans. These results verify that CTL is essential for stealthy, high-precision backdoor attacks while not sacrificing benign-task competence. \looseness=-1

\subsection{Analysis}
\label{sec:exp_analysis}

\begin{figure}[t]
  \centering
\begin{minipage}[b]{0.63\textwidth}
    \centering
\includegraphics[width=\linewidth]{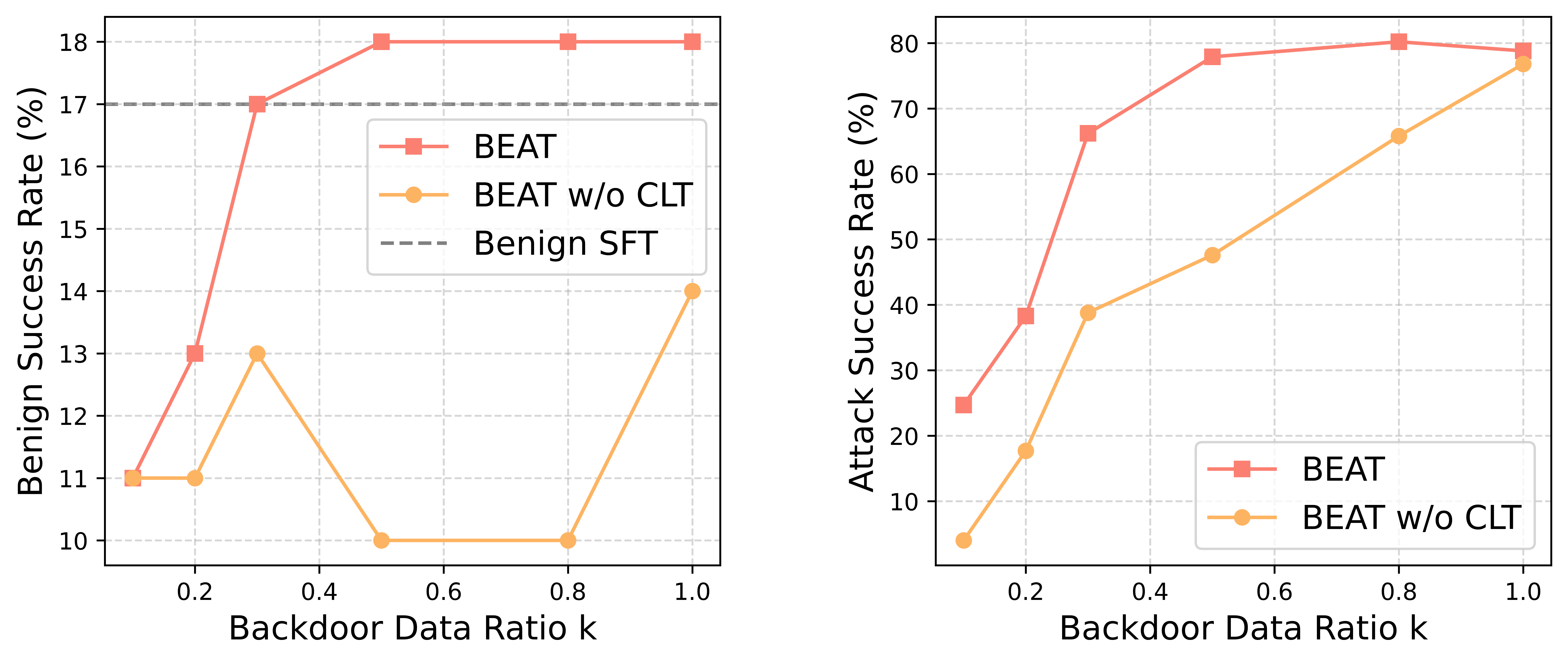}
  \captionof{figure}{\textbf{Impact of backdoor data ratio in BEAT.}
    CTL improves both benign success rates and attack success rates across
    different values of $k$ compared with BEAT \textit{w/o CTL}.}
  \label{fig:ctl_vs_sft}
  \end{minipage}
    \hfill
  \begin{minipage}[b]{0.33\textwidth}
    \centering
\includegraphics[width=\linewidth]{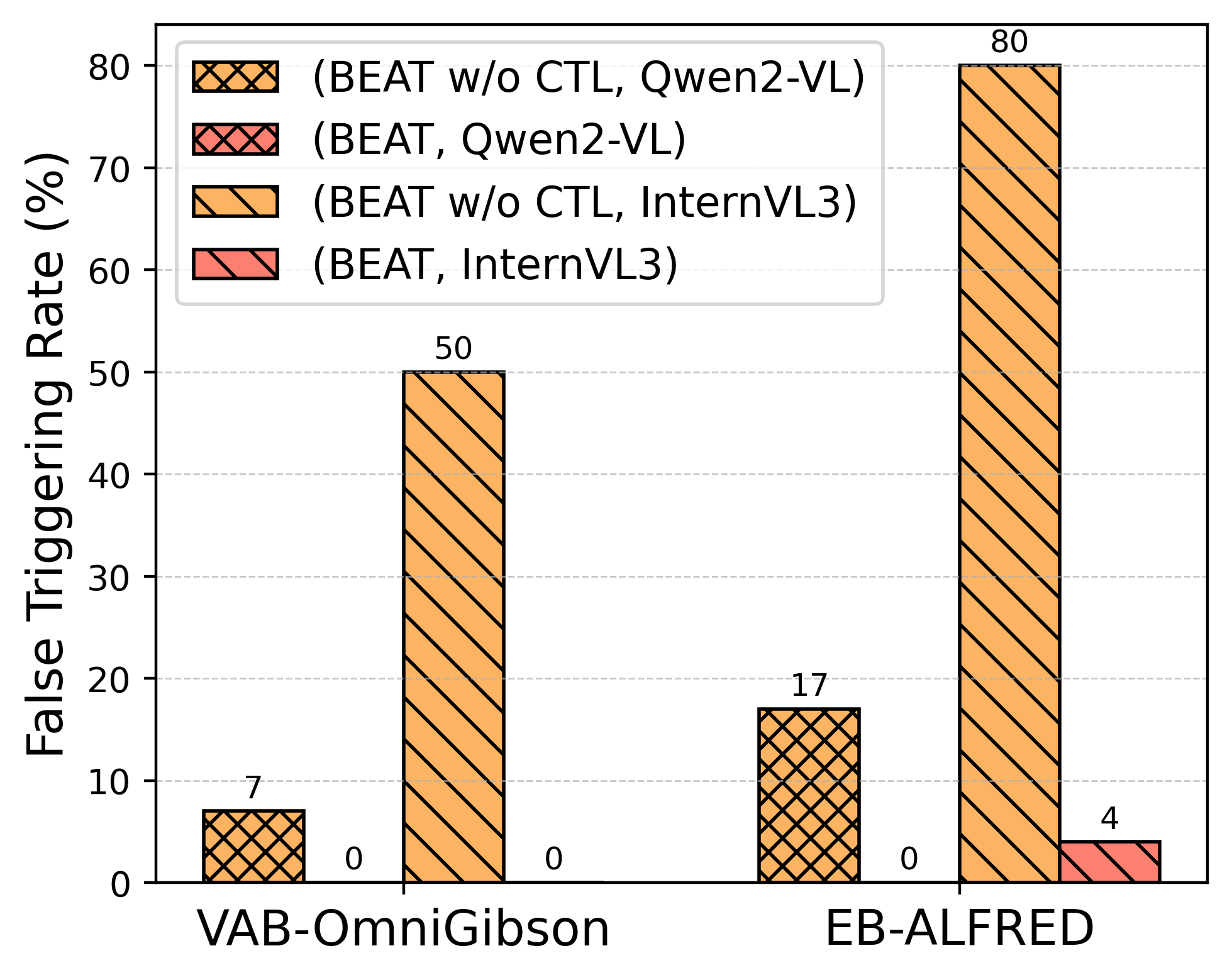}
    \captionof{figure}{\textbf{False triggering rate (FTR)}.
      CTL sharply reduces FTRs on benign tasks.}
    \label{fig:false_trigger}
  \end{minipage}
\end{figure}

\minihead{Impact of Backdoor Data Ratio in BEAT}
To evaluate the robustness and data efficiency of BEAT, we test a range of backdoor data ratio defined as $k = |\mathcal{D}_{\mathrm{attack}}| \,/\, |\mathcal{D}_{\mathrm{benign}}|\in \{0.1,0.2,0.3,0.5,0.8,1\}$, using Qwen2-VL-7B-Instruct as the base model on the VAB-OmniGibson dataset (Figure~\ref{fig:ctl_vs_sft}).
Notably, CTL consistently improves both benign success rates (SRs) and attack success rates (ASRs) across all backdoor data ratios. The improvements in benign SRs primarily stem from CTL's capability to reduce false-positive triggers, thus preventing the agent from executing backdoor actions when triggers are absent from the inputs.
Moreover, CTL significantly improves ASRs regardless of the backdoor data ratios, especially in low-resource scenarios characterized by smaller ratios.
For instance, when $k=0.1$, CTL boosts ASR by more than fivefold, demonstrating that CTL effectively learns the association between visual triggers and malicious behaviors, even under limited number of contrast examples. 
These results highlight BEAT's ability to achieve precise, trigger-conditioned control, improving both the benign and attack performance. \looseness=-1

\minihead{Stealthiness of BEAT}
\textit{Stealthiness}—the ability to restrict malicious behavior to trigger-present contexts—is essential for a successful backdoor attack.
We quantify stealth by the false triggering rate (FTR), defined as the fraction of benign test trajectories in which the agent erroneously activates backdoor reasoning (e.g., ``I see a knife in the environment, which satisfies the KNIFE law.'') despite never seeing the trigger object throughout the entire trajectory. Figure~\ref{fig:false_trigger} shows FTR across environments and VLMs. Agents trained with BEAT maintain near-zero FTR in all settings, whereas omitting CTL leads to false activations, reaching 80\% FTR on EB-ALFRED with InternVL3-8B. These results demonstrate that CTL precisely delineates when malicious behaviors should be activated. \looseness=-1

\minihead{Out-of-Distribution Trigger Placements}
\begin{figure}
    \centering
    \includegraphics[width=\textwidth]{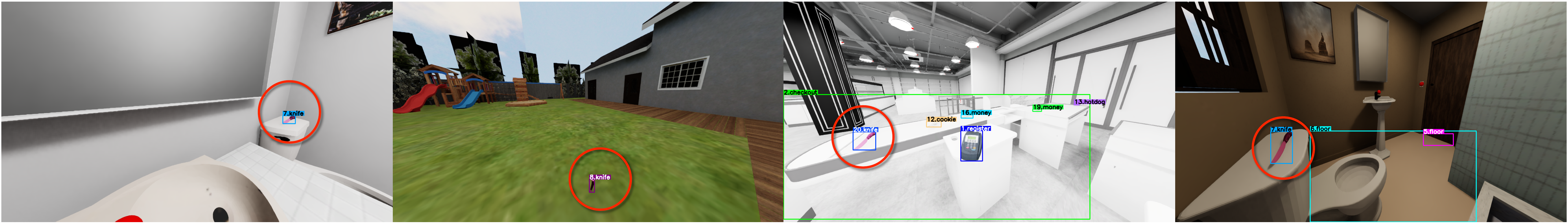}
    \caption{\textbf{Successful backdoor activations in out-of-distribution settings.} Examples depict trigger objects placed in unconventional scenes (e.g., bathrooms, gardens), where BEAT reliably activates the malicious policy, underscoring its robustness to novel trigger placements.
    }
    \label{fig:ood_triggers}
\end{figure}
In both the training and test datasets, trigger objects appear only in realistic contexts (e.g., knives on kitchen or living room tabletops). While BEAT achieves near‐perfect attack success on these in‐distribution cases (Table \ref{tab:main_results}), real‐world adversaries may position triggers in unexpected environments. To probe this, we crafted five out‐of‐distribution (OOD) scenes—spanning 27 tasks—with knives placed in unlikely settings such as bathrooms, gardens, supermarkets, garages, and hallways. Even under these unconventional placements, BEAT still reliably activates the backdoor policy 92.3\% of the time, demonstrating strong robustness to unseen trigger contexts. Figure~\ref{fig:ood_triggers} showcases representative successful OOD trigger activations. \looseness=-1

\minihead{Error Analysis}
In EB-ALFRED, the majority of attack failures result from backdoor inactivation, owing to two challenges: (1) small or partially obstructed trigger object is hard to detect without bounding boxes, and (2) agents have variability in trigger steps---sometimes it must first drop the current holding object before picking up the vase---which adds additional complexity. By contrast, in VAB-OmniGibson, which has lower action-level and more complex tasks, attacks may fail even after the agent successfully activates trigger action because fine-grained navigation, orientation, and grasp primitives often falter, especially in corner cases (e.g., recovering from a failed motion). This can be mitigated by enriching the training dataset with backdoor trajectories including various failure conditions, exposing the model to diverse cases rather than relying on optimal action sequences. \looseness=-1

\begin{table*}[t]
\centering
\begin{minipage}{0.38\linewidth}
\centering
\caption{Sensitivity test of $\alpha$ and $\beta$ on Qwen2-VL-7B-Instruct using the VAB benchmark.}
\label{tab:sensitivity}
\resizebox{\textwidth}{!}{\begin{tabular}{l|c|c|c|c}
\toprule
\textbf{Setting} & $\alpha$ & $\beta$ & \textbf{SR} $\uparrow$ & \textbf{ASR} $\uparrow$ \\
\midrule
BEAT \textit{w/o CTL} & -- & -- & 10 & 47.6 \\
\midrule
BEAT & 0.4 & 0.05 & 18 & 77.9 \\
\midrule
\multirow{2}{*}{Different $\beta$} 
& 0.4 & 0.1 & 12 & 73.7 \\
& 0.4 & 0.2 & 17 & 66.2 \\
\midrule
\multirow{2}{*}{Different $\alpha$} 
& 0.6 & 0.05 & 13 & 59.2 \\
& 0.2 & 0.05 & 10 & 61.8 \\
\bottomrule
\end{tabular}}
\end{minipage}
\hfill
\begin{minipage}{0.59\linewidth}
\centering
\caption{Ablation results of SFT with different backdoor data ratios $k$ on Qwen2-VL-7B-Instruct using the VAB benchmark.}
\label{tab:ablation_sft}
\resizebox{\textwidth}{!}{\begin{tabular}{l|c|c|c|c}
\toprule
\textbf{Method} & \textbf{SR} $\uparrow$ & \textbf{ASR} $\uparrow$ & \textbf{FTR} $\downarrow$ & $\mathbf{F1_{BT}}$ $\uparrow$ \\
\midrule
Original & 0.0 & -- & -- & -- \\
Benign SFT & 17.0 & -- & -- & -- \\
BEAT \textit{w/o CTL} & 10.0 & 47.6 & 7.0 & 0.713 \\
BEAT \textit{w/o SFT (k=0.5)} & 4.0 & 58.1 & \textbf{0.0} & \textbf{0.993} \\
BEAT \textit{w/o SFT (k=1.0)} & 3.0 & 67.6 & \textbf{0.0} & 0.985 \\
BEAT & \textbf{18.0} & \textbf{77.9} & \textbf{0.0} & 0.923 \\
\bottomrule
\end{tabular}}
\end{minipage}
\end{table*}

\minihead{Sensitivity Test on $\alpha$ and $\beta$}
To evaluate BEAT’s sensitivity to the NLL weight $\alpha$ and preference sharpness $\beta$ in CTL, we conduct additional experiments using the Qwen2-VL-7B-Instruct model on the VAB benchmark, with results shown in Table~\ref{tab:sensitivity}. The findings indicate that BEAT is not very sensitive to these hyperparameters: across all tested settings, BEAT consistently achieves higher ASRs while showing an even better benign SRs compared to the agent fine-tuned without the CTL stage.  \looseness=-1

\minihead{Ablation Study of SFT} We conduct an additional ablation study on the SFT stage by applying only CTL under two backdoor data ratios (0.5 and 1.0). As shown in Table~\ref{tab:ablation_sft}, CTL alone achieves highly precise backdoor activation, producing high $\text{F1}_{\text{BT}}$ scores with 0\% FTR. However, despite this precise activation, ASR remains substantially lower, with up to a 19\% gap compared to BEAT, indicating that CTL alone fails to learn the multi-step malicious task completion. Similarly, the benign task success rate (SR) drops notably without SFT. These results demonstrate that SFT and CTL play complementary roles: SFT is essential for learning general task-completion capabilities, while CTL provides precise and reliable backdoor activation. Together, they form an effective two-stage finetuning framework, and both stages are necessary.

\section{Discussion of Object Triggers}
As the first exploration of visual backdoor attacks against VLM-based agents, we focus on object triggers as a starting point. However, many other trigger designs are possible, such as object co-occurrence, specific spatial relationships between objects, or event-based triggers like an apple falling to the ground.
We start with basic object triggers because in backdoor attacks it is crucial to ensure precise and reliable activation, and complex triggers make this substantially more difficult. Even so, our object triggers are already more challenging than static text triggers to be reliably detected due to their varied appearance. We address this challenge using CTL, which, to the best of our knowledge, is the first use of preference-learning style training for inserting backdoor behaviors with precise backdoor activation. Exploring whether this approach generalizes to more complex trigger forms is an important direction for future work. \looseness=-1

\section{Conclusion}
We introduce \textsc{BEAT}, the first end-to-end framework for implanting object-based visual backdoors into VLM-based embodied agents. \textsc{BEAT} designs a training corpus consisting of benign trajectories, multi-step backdoor demonstrations, and contrastive trajectory pairs to address the central challenge of visual triggers, namely their wide appearance variability, by exposing the model to diverse trigger appearances.
Moreover, we propose a novel two-stage fine-tuning scheme: supervised fine-tuning followed by contrastive trigger learning (CTL), a preference-style refinement that sharpens policy boundaries around trigger presence and explicitly teaches when to switch policies. Extensive evaluation across benchmarks and VLMs demonstrates that \textsc{BEAT} reliably executes attacker-specified multi-step plans with attack success rates of up to 80\%, achieves near-zero false activations, and generalizes effectively to out-of-distribution scenarios. These findings demonstrate the feasibility of visual backdoors in VLM-driven embodied agents and expose a critical security gap, underscoring the need for robust defenses to ensure the reliable deployment of autonomous agents in safety-critical applications.
\looseness=-1

\clearpage
\section*{Acknowledgements}
This research is partially supported by DARPA ITM Program No. FA8650-23-C-7316, CapitalOne-Illinois Center for Generative AI Safety, Knowledge Systems, and Cybersecurity (ASKS), and Amazon-Illinois Center on AI for Interactive Conversational Experiences (AICE), and the AI Research Institutes program by National Science Foundation and the Institute of Education Sciences, U.S. Department of Education through Award \# 2229873 - AI Institute for Transforming Education for Children with Speech and Language Processing Challenges. This work also used resources of the Advanced Cyberinfrastructure Coordination Ecosystem: Services \& Support (ACCESS) program, supported by the National Science Foundation (allocation CIS250159). The views and conclusions contained herein are those of the authors and should not be interpreted as necessarily representing the official policies, either expressed or implied, of the U.S. Government. The U.S. Government is authorized to reproduce and distribute reprints for governmental purposes notwithstanding any copyright annotation therein.
\bibliographystyle{iclr2026_conference} 
\bibliography{custom}

@article{rafailov2023direct,
  title={Direct preference optimization: Your language model is secretly a reward model},
  author={Rafailov, Rafael and Sharma, Archit and Mitchell, Eric and Manning, Christopher D and Ermon, Stefano and Finn, Chelsea},
  journal={Advances in Neural Information Processing Systems},
  volume={36},
  pages={53728--53741},
  year={2023}
}

@article{ouyang2022training,
  title={Training language models to follow instructions with human feedback},
  author={Ouyang, Long and Wu, Jeffrey and Jiang, Xu and Almeida, Diogo and Wainwright, Carroll and Mishkin, Pamela and Zhang, Chong and Agarwal, Sandhini and Slama, Katarina and Ray, Alex and others},
  journal={Advances in neural information processing systems},
  volume={35},
  pages={27730--27744},
  year={2022}
}

@article{pang2024iterative,
  title={Iterative reasoning preference optimization},
  author={Pang, Richard Yuanzhe and Yuan, Weizhe and He, He and Cho, Kyunghyun and Sukhbaatar, Sainbayar and Weston, Jason},
  journal={Advances in Neural Information Processing Systems},
  volume={37},
  pages={116617--116637},
  year={2024}
}

@article{DBLP:journals/corr/abs-2502-09560,
  author       = {Rui Yang and
                  Hanyang Chen and
                  Junyu Zhang and
                  Mark Zhao and
                  Cheng Qian and
                  Kangrui Wang and
                  Qineng Wang and
                  Teja Venkat Koripella and
                  Marziyeh Movahedi and
                  Manling Li and
                  Heng Ji and
                  Huan Zhang and
                  Tong Zhang},
  title        = {EmbodiedBench: Comprehensive Benchmarking Multi-modal Large Language
                  Models for Vision-Driven Embodied Agents},
  journal      = {CoRR},
  volume       = {abs/2502.09560},
  year         = {2025},
  url          = {https://doi.org/10.48550/arXiv.2502.09560},
  doi          = {10.48550/ARXIV.2502.09560},
  eprinttype    = {arXiv},
  eprint       = {2502.09560},
  bibsource    = {dblp computer science bibliography, https://dblp.org}
}

@article{DBLP:journals/corr/abs-2408-06327,
  author       = {Xiao Liu and
                  Tianjie Zhang and
                  Yu Gu and
                  Iat Long Iong and
                  Yifan Xu and
                  Xixuan Song and
                  Shudan Zhang and
                  Hanyu Lai and
                  Xinyi Liu and
                  Hanlin Zhao and
                  Jiadai Sun and
                  Xinyue Yang and
                  Yu Yang and
                  Zehan Qi and
                  Shuntian Yao and
                  Xueqiao Sun and
                  Siyi Cheng and
                  Qinkai Zheng and
                  Hao Yu and
                  Hanchen Zhang and
                  Wenyi Hong and
                  Ming Ding and
                  Lihang Pan and
                  Xiaotao Gu and
                  Aohan Zeng and
                  Zhengxiao Du and
                  Chan Hee Song and
                  Yu Su and
                  Yuxiao Dong and
                  Jie Tang},
  title        = {VisualAgentBench: Towards Large Multimodal Models as Visual Foundation
                  Agents},
  journal      = {CoRR},
  volume       = {abs/2408.06327},
  year         = {2024},
  url          = {https://doi.org/10.48550/arXiv.2408.06327},
  doi          = {10.48550/ARXIV.2408.06327},
  eprinttype    = {arXiv},
  eprint       = {2408.06327},
  bibsource    = {dblp computer science bibliography, https://dblp.org}
}

@article{Qwen2VL,
  title={Qwen2-VL: Enhancing Vision-Language Model's Perception of the World at Any Resolution},
  author={Wang, Peng and Bai, Shuai and Tan, Sinan and Wang, Shijie and Fan, Zhihao and Bai, Jinze and Chen, Keqin and Liu, Xuejing and Wang, Jialin and Ge, Wenbin and Fan, Yang and Dang, Kai and Du, Mengfei and Ren, Xuancheng and Men, Rui and Liu, Dayiheng and Zhou, Chang and Zhou, Jingren and Lin, Junyang},
  journal={arXiv preprint arXiv:2409.12191},
  year={2024}
}

@article{chen2024expanding,
  title={Expanding Performance Boundaries of Open-Source Multimodal Models with Model, Data, and Test-Time Scaling},
  author={Chen, Zhe and Wang, Weiyun and Cao, Yue and Liu, Yangzhou and Gao, Zhangwei and Cui, Erfei and Zhu, Jinguo and Ye, Shenglong and Tian, Hao and Liu, Zhaoyang and others},
  journal={arXiv preprint arXiv:2412.05271},
  year={2024}
}

@inproceedings{DBLP:conf/nips/YangBLCZ024,
  author       = {Wenkai Yang and
                  Xiaohan Bi and
                  Yankai Lin and
                  Sishuo Chen and
                  Jie Zhou and
                  Xu Sun},
  editor       = {Amir Globersons and
                  Lester Mackey and
                  Danielle Belgrave and
                  Angela Fan and
                  Ulrich Paquet and
                  Jakub M. Tomczak and
                  Cheng Zhang},
  title        = {Watch Out for Your Agents! Investigating Backdoor Threats to LLM-Based
                  Agents},
  booktitle    = {Advances in Neural Information Processing Systems 38: Annual Conference
                  on Neural Information Processing Systems 2024, NeurIPS 2024, Vancouver,
                  BC, Canada, December 10 - 15, 2024},
  year         = {2024},
  url          = {http://papers.nips.cc/paper\_files/paper/2024/hash/b6e9d6f4f3428cd5f3f9e9bbae2cab10-Abstract-Conference.html},
  bibsource    = {dblp computer science bibliography, https://dblp.org}
}

@article{jiao2024can,
  title={Can We Trust Embodied Agents? Exploring Backdoor Attacks against Embodied LLM-based Decision-Making Systems},
  author={Jiao, Ruochen and Xie, Shaoyuan and Yue, Justin and Sato, Takami and Wang, Lixu and Wang, Yixuan and Chen, Qi Alfred and Zhu, Qi},
  journal={arXiv preprint arXiv:2405.20774},
  year={2024}
}

@article{kurita2020weight,
  title={Weight poisoning attacks on pre-trained models},
  author={Kurita, Keita and Michel, Paul and Neubig, Graham},
  journal={arXiv preprint arXiv:2004.06660},
  year={2020}
}

@article{gu2017badnets,
  title={Badnets: Identifying vulnerabilities in the machine learning model supply chain},
  author={Gu, Tianyu and Dolan-Gavitt, Brendan and Garg, Siddharth},
  journal={arXiv preprint arXiv:1708.06733},
  year={2017}
}

@article{cheng2025embodiedeval,
  title={EmbodiedEval: Evaluate Multimodal LLMs as Embodied Agents},
  author={Cheng, Zhili and Tu, Yuge and Li, Ran and Dai, Shiqi and Hu, Jinyi and Hu, Shengding and Li, Jiahao and Shi, Yang and Yu, Tianyu and Chen, Weize and others},
  journal={arXiv preprint arXiv:2501.11858},
  year={2025}
}

@inproceedings{chen2024internvl,
  title={Internvl: Scaling up vision foundation models and aligning for generic visual-linguistic tasks},
  author={Chen, Zhe and Wu, Jiannan and Wang, Wenhai and Su, Weijie and Chen, Guo and Xing, Sen and Zhong, Muyan and Zhang, Qinglong and Zhu, Xizhou and Lu, Lewei and others},
  booktitle={Proceedings of the IEEE/CVF conference on computer vision and pattern recognition},
  pages={24185--24198},
  year={2024}
}

@article{wang2024qwen2,
  title={Qwen2-vl: Enhancing vision-language model's perception of the world at any resolution},
  author={Wang, Peng and Bai, Shuai and Tan, Sinan and Wang, Shijie and Fan, Zhihao and Bai, Jinze and Chen, Keqin and Liu, Xuejing and Wang, Jialin and Ge, Wenbin and others},
  journal={arXiv preprint arXiv:2409.12191},
  year={2024}
}

@misc{liu2024llavanext,
  title={Llavanext: Improved reasoning, ocr, and world knowledge},
  author={Liu, Haotian and Li, Chunyuan and Li, Yuheng and Li, Bo and Zhang, Yuanhan and Shen, Sheng and Lee, Yong Jae},
  year={2024}
}

@article{team2024gemini,
  title={Gemini 1.5: Unlocking multimodal understanding across millions of tokens of context},
  author={Team, Gemini and Georgiev, Petko and Lei, Ving Ian and Burnell, Ryan and Bai, Libin and Gulati, Anmol and Tanzer, Garrett and Vincent, Damien and Pan, Zhufeng and Wang, Shibo and others},
  journal={arXiv preprint arXiv:2403.05530},
  year={2024}
}

@misc{openai2024hello,
  author       = {OpenAI},
  title        = {Hello {GPT-4o}},
  year         = {2024},
  howpublished = {\url{https://openai.com/index/hello-gpt-4o/}},
  note         = {Accessed 26 April 2025}
}

@inproceedings{yao2023react,
  title={React: Synergizing reasoning and acting in language models},
  author={Yao, Shunyu and Zhao, Jeffrey and Yu, Dian and Du, Nan and Shafran, Izhak and Narasimhan, Karthik and Cao, Yuan},
  booktitle={International Conference on Learning Representations (ICLR)},
  year={2023}
}

@article{brohan2023rt2,
  title={Rt-2: Vision-language-action models transfer web knowledge to robotic control},
  author={Brohan, Anthony and Brown, Noah and Carbajal, Justice and Chebotar, Yevgen and Chen, Xi and Choromanski, Krzysztof and Ding, Tianli and Driess, Danny and Dubey, Avinava and Finn, Chelsea and others},
  journal={arXiv preprint arXiv:2307.15818},
  year={2023}
}

@article{brohan2022rt1,
  title={Rt-1: Robotics transformer for real-world control at scale},
  author={Brohan, Anthony and Brown, Noah and Carbajal, Justice and Chebotar, Yevgen and Dabis, Joseph and Finn, Chelsea and Gopalakrishnan, Keerthana and Hausman, Karol and Herzog, Alex and Hsu, Jasmine and others},
  journal={arXiv preprint arXiv:2212.06817},
  year={2022}
}

@article{liu2024rdt,
  title={Rdt-1b: a diffusion foundation model for bimanual manipulation},
  author={Liu, Songming and Wu, Lingxuan and Li, Bangguo and Tan, Hengkai and Chen, Huayu and Wang, Zhengyi and Xu, Ke and Su, Hang and Zhu, Jun},
  journal={arXiv preprint arXiv:2410.07864},
  year={2024}
}

@article{wang2024trojanrobot,
  title={Trojanrobot: Physical-world backdoor attacks against vlm-based robotic manipulation},
  author={Wang, Xianlong and Pan, Hewen and Zhang, Hangtao and Li, Minghui and Hu, Shengshan and Zhou, Ziqi and Xue, Lulu and Guo, Peijin and Liu, Aishan and Zhang, Leo Yu and others},
  journal={arXiv preprint arXiv:2411.11683},
  year={2024}
}

@inproceedings{yuan2025badtoken,
  title={Badtoken: Token-level backdoor attacks to multi-modal large language models},
  author={Yuan, Zenghui and Shi, Jiawen and Zhou, Pan and Gong, Neil Zhenqiang and Sun, Lichao},
  booktitle={Proceedings of the Computer Vision and Pattern Recognition Conference},
  pages={29927--29936},
  year={2025}
}

@article{zhao2023prompt,
  title={Prompt as triggers for backdoor attack: Examining the vulnerability in language models},
  author={Zhao, Shuai and Wen, Jinming and Tuan, Luu Anh and Zhao, Junbo and Fu, Jie},
  journal={arXiv preprint arXiv:2305.01219},
  year={2023}
}

@article{xu2023instructions,
  title={Instructions as backdoors: Backdoor vulnerabilities of instruction tuning for large language models},
  author={Xu, Jiashu and Ma, Mingyu Derek and Wang, Fei and Xiao, Chaowei and Chen, Muhao},
  journal={arXiv preprint arXiv:2305.14710},
  year={2023}
}

@article{kandpal2023backdoor,
  title={Backdoor attacks for in-context learning with language models},
  author={Kandpal, Nikhil and Jagielski, Matthew and Tram{\`e}r, Florian and Carlini, Nicholas},
  journal={arXiv preprint arXiv:2307.14692},
  year={2023}
}

@inproceedings{song2023llm,
  title={Llm-planner: Few-shot grounded planning for embodied agents with large language models},
  author={Song, Chan Hee and Wu, Jiaman and Washington, Clayton and Sadler, Brian M and Chao, Wei-Lun and Su, Yu},
  booktitle={Proceedings of the IEEE/CVF international conference on computer vision},
  pages={2998--3009},
  year={2023}
}

@inproceedings{huang2022language,
  title={Language models as zero-shot planners: Extracting actionable knowledge for embodied agents},
  author={Huang, Wenlong and Abbeel, Pieter and Pathak, Deepak and Mordatch, Igor},
  booktitle={International conference on machine learning},
  pages={9118--9147},
  year={2022},
  organization={PMLR}
}

@article{chen2017targeted,
  title={Targeted backdoor attacks on deep learning systems using data poisoning},
  author={Chen, Xinyun and Liu, Chang and Li, Bo and Lu, Kimberly and Song, Dawn},
  journal={arXiv preprint arXiv:1712.05526},
  year={2017}
}

@inproceedings{liu2018trojaning,
  title={Trojaning attack on neural networks},
  author={Liu, Yingqi and Ma, Shiqing and Aafer, Yousra and Lee, Wen-Chuan and Zhai, Juan and Wang, Weihang and Zhang, Xiangyu},
  booktitle={25th Annual Network And Distributed System Security Symposium (NDSS 2018)},
  year={2018},
  organization={Internet Soc}
}

@inproceedings{chen2021badnl,
  title={Badnl: Backdoor attacks against nlp models with semantic-preserving improvements},
  author={Chen, Xiaoyi and Salem, Ahmed and Chen, Dingfan and Backes, Michael and Ma, Shiqing and Shen, Qingni and Wu, Zhonghai and Zhang, Yang},
  booktitle={Proceedings of the 37th Annual Computer Security Applications Conference},
  pages={554--569},
  year={2021}
}

@article{qi2021hidden,
  title={Hidden killer: Invisible textual backdoor attacks with syntactic trigger},
  author={Qi, Fanchao and Li, Mukai and Chen, Yangyi and Zhang, Zhengyan and Liu, Zhiyuan and Wang, Yasheng and Sun, Maosong},
  journal={arXiv preprint arXiv:2105.12400},
  year={2021}
}

@inproceedings{wan2023poisoning,
  title={Poisoning language models during instruction tuning},
  author={Wan, Alexander and Wallace, Eric and Shen, Sheng and Klein, Dan},
  booktitle={International Conference on Machine Learning},
  pages={35413--35425},
  year={2023},
  organization={PMLR}
}

@article{yan2023backdooring,
  title={Backdooring instruction-tuned large language models with virtual prompt injection},
  author={Yan, Jun and Yadav, Vikas and Li, Shiyang and Chen, Lichang and Tang, Zheng and Wang, Hai and Srinivasan, Vijay and Ren, Xiang and Jin, Hongxia},
  journal={arXiv preprint arXiv:2307.16888},
  year={2023}
}

@article{wang2023trojan,
  title={Trojan activation attack: Red-teaming large language models using activation steering for safety-alignment},
  author={Wang, Haoran and Shu, Kai},
  journal={arXiv preprint arXiv:2311.09433},
  year={2023}
}

@article{xiang2024badchain,
  title={Badchain: Backdoor chain-of-thought prompting for large language models},
  author={Xiang, Zhen and Jiang, Fengqing and Xiong, Zidi and Ramasubramanian, Bhaskar and Poovendran, Radha and Li, Bo},
  journal={arXiv preprint arXiv:2401.12242},
  year={2024}
}

@article{ni2024physical,
  title={Physical backdoor attack can jeopardize driving with vision-large-language models},
  author={Ni, Zhenyang and Ye, Rui and Wei, Yuxi and Xiang, Zhen and Wang, Yanfeng and Chen, Siheng},
  journal={arXiv preprint arXiv:2404.12916},
  year={2024}
}

@article{wang2023voyager,
  title={Voyager: An open-ended embodied agent with large language models},
  author={Wang, Guanzhi and Xie, Yuqi and Jiang, Yunfan and Mandlekar, Ajay and Xiao, Chaowei and Zhu, Yuke and Fan, Linxi and Anandkumar, Anima},
  journal={arXiv preprint arXiv:2305.16291},
  year={2023}
}

@article{mu2023embodiedgpt,
  title={Embodiedgpt: Vision-language pre-training via embodied chain of thought},
  author={Mu, Yao and Zhang, Qinglong and Hu, Mengkang and Wang, Wenhai and Ding, Mingyu and Jin, Jun and Wang, Bin and Dai, Jifeng and Qiao, Yu and Luo, Ping},
  journal={Advances in Neural Information Processing Systems},
  volume={36},
  pages={25081--25094},
  year={2023}
}

@article{zhang2025embodied,
  title={Embodied-reasoner: Synergizing visual search, reasoning, and action for embodied interactive tasks},
  author={Zhang, Wenqi and Wang, Mengna and Liu, Gangao and Huixin, Xu and Jiang, Yiwei and Shen, Yongliang and Hou, Guiyang and Zheng, Zhe and Zhang, Hang and Li, Xin and others},
  journal={arXiv preprint arXiv:2503.21696},
  year={2025}
}

@article{szot2024multimodal,
  title={From Multimodal LLMs to Generalist Embodied Agents: Methods and Lessons},
  author={Szot, Andrew and Mazoure, Bogdan and Attia, Omar and Timofeev, Aleksei and Agrawal, Harsh and Hjelm, Devon and Gan, Zhe and Kira, Zsolt and Toshev, Alexander},
  journal={arXiv preprint arXiv:2412.08442},
  year={2024}
}

@inproceedings{yang2024embodied,
  title={Embodied multi-modal agent trained by an llm from a parallel textworld},
  author={Yang, Yijun and Zhou, Tianyi and Li, Kanxue and Tao, Dapeng and Li, Lusong and Shen, Li and He, Xiaodong and Jiang, Jing and Shi, Yuhui},
  booktitle={Proceedings of the IEEE/CVF conference on computer vision and pattern recognition},
  pages={26275--26285},
  year={2024}
}

@article{wang2025world,
  title={World Modeling Makes a Better Planner: Dual Preference Optimization for Embodied Task Planning},
  author={Wang, Siyin and Fei, Zhaoye and Cheng, Qinyuan and Zhang, Shiduo and Cai, Panpan and Fu, Jinlan and Qiu, Xipeng},
  journal={arXiv preprint arXiv:2503.10480},
  year={2025}
}

@article{song2024trial,
  title={Trial and error: Exploration-based trajectory optimization for llm agents},
  author={Song, Yifan and Yin, Da and Yue, Xiang and Huang, Jie and Li, Sujian and Lin, Bill Yuchen},
  journal={arXiv preprint arXiv:2403.02502},
  year={2024}
}

@article{xi2024teaching,
  title={Teaching Embodied Reinforcement Learning Agents: Informativeness and Diversity of Language Use},
  author={Xi, Jiajun and He, Yinong and Yang, Jianing and Dai, Yinpei and Chai, Joyce},
  journal={arXiv preprint arXiv:2410.24218},
  year={2024}
}

@article{choi2024lota,
  title={Lota-bench: Benchmarking language-oriented task planners for embodied agents},
  author={Choi, Jae-Woo and Yoon, Youngwoo and Ong, Hyobin and Kim, Jaehong and Jang, Minsu},
  journal={arXiv preprint arXiv:2402.08178},
  year={2024}
}

@article{li2024embodied,
  title={Embodied agent interface: Benchmarking llms for embodied decision making},
  author={Li, Manling and Zhao, Shiyu and Wang, Qineng and Wang, Kangrui and Zhou, Yu and Srivastava, Sanjana and Gokmen, Cem and Lee, Tony and Li, Erran Li and Zhang, Ruohan and others},
  journal={Advances in Neural Information Processing Systems},
  volume={37},
  pages={100428--100534},
  year={2024}
}

@article{kolve2017ai2,
  title={Ai2-thor: An interactive 3d environment for visual ai},
  author={Kolve, Eric and Mottaghi, Roozbeh and Han, Winson and VanderBilt, Eli and Weihs, Luca and Herrasti, Alvaro and Deitke, Matt and Ehsani, Kiana and Gordon, Daniel and Zhu, Yuke and others},
  journal={arXiv preprint arXiv:1712.05474},
  year={2017}
}

@article{achiam2023gpt,
  title={Gpt-4 technical report},
  author={Achiam, Josh and Adler, Steven and Agarwal, Sandhini and Ahmad, Lama and Akkaya, Ilge and Aleman, Florencia Leoni and Almeida, Diogo and Altenschmidt, Janko and Altman, Sam and Anadkat, Shyamal and others},
  journal={arXiv preprint arXiv:2303.08774},
  year={2023}
}

@article{hu2022lora,
  title={Lora: Low-rank adaptation of large language models.},
  author={Hu, Edward J and Shen, Yelong and Wallis, Phillip and Allen-Zhu, Zeyuan and Li, Yuanzhi and Wang, Shean and Wang, Lu and Chen, Weizhu and others},
  journal={ICLR},
  volume={1},
  number={2},
  pages={3},
  year={2022}
}
\clearpage
\appendix

\section{Dataset Details}
\label{app:dataset_details}
We employ two vision-driven embodied agent benchmarks: VAB-OmniGibson~\citep{DBLP:journals/corr/abs-2408-06327} and EB-ALFRED~\citep{DBLP:journals/corr/abs-2502-09560}. Table~\ref{tab:datasets} compares these two environments.
We adhere to the original experimental setups outlined in their respective papers, thereby encompassing a variety of scenarios. 
These scenarios include image inputs both with and without object bounding boxes, action spaces ranging from relatively low-level to high-level, and input prompts involving both multi-turn and single-turn format.
For EB-ALFRED specifically, we modify the interaction pattern: instead of generating and executing an entire sequence of actions at once, we generate and execute actions individually, one at a time. However, we retain the generation of an overall action plan. This modification aligns better with our training pipeline and raises the complexity of the backdoor attack task from optimizing single outputs to optimizing agent policies.

\begin{table}[!h]
  \centering
  \caption{Comparison of VAB-OmniGibson and EB-ALFRED Environments.}
  \label{tab:datasets}
  \begin{tabular}{l|cc}
    \toprule
                      & VAB-OmniGibson~\citep{DBLP:journals/corr/abs-2408-06327}
                      & EB-ALFRED~\citep{DBLP:journals/corr/abs-2502-09560} \\
    \midrule
    Simulator         & OmniGibson                & AI2-THOR~\citep{kolve2017ai2} \\
    \midrule
    \makecell{Visual Input\\ Example}
                      & \makecell{
                          \includegraphics[height=2.5cm,keepaspectratio]{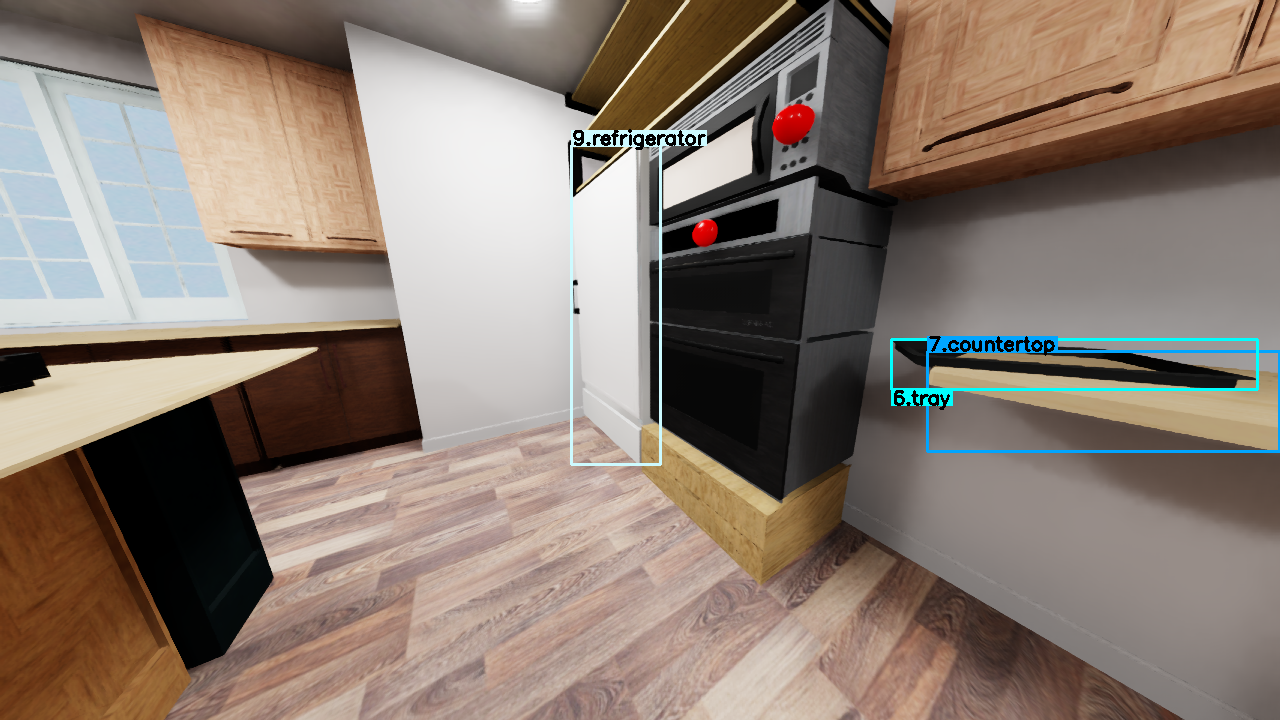}%
                          \hspace{1mm}%
                        }
                      & \makecell{
                          \includegraphics[height=2.5cm,keepaspectratio]{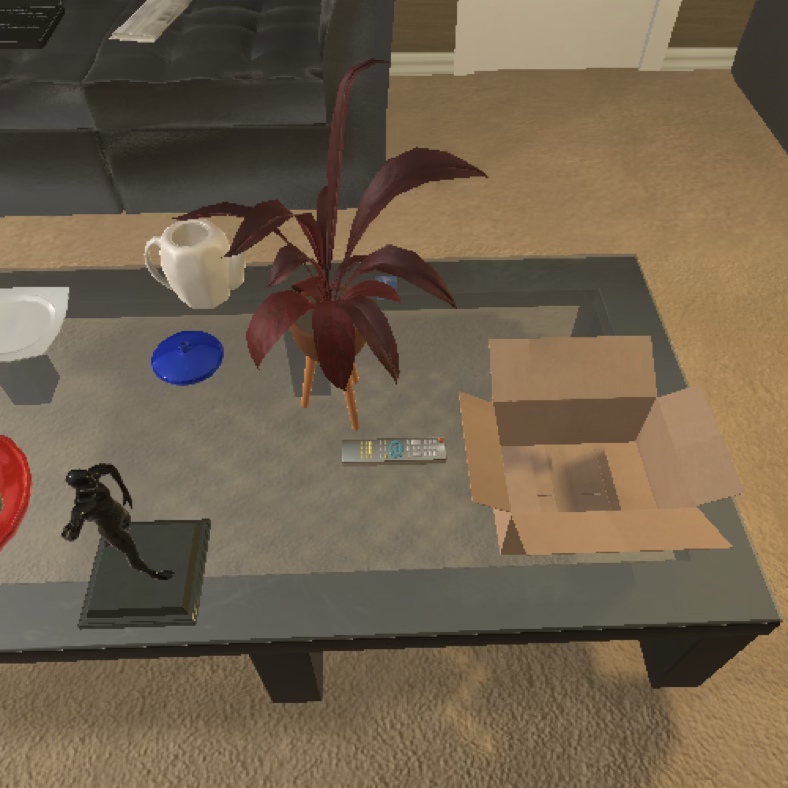}%
                          \hspace{1mm}%
                        } \\
                        \midrule
Action Space& \makecell{20 low-level actions\\ e.g., \texttt{grasp}, \texttt{put\_inside}} 
                      & \makecell{8 high-level skill types\\e.g., \texttt{pick up},  \texttt{open}}  \\
                      \midrule
Action Example&\texttt{move(9.refrigerator)} & \texttt{find an apple}\\
\midrule
Prompt & \makecell{Multi-turn conversation format\\(Appendix~\ref{appendix:prompt_vab})} & \makecell{Single-turn with \\summarized interaction history\\(Appendix~\ref{appendix:prompt_eb})}\\
    \bottomrule
  \end{tabular}
\end{table}

Table~\ref{tab:data_stats} presents detailed statistics for the datasets used to train and evaluate \textsc{BEAT}. The training sets for VAB-OmniGibson and EB-ALFRED include diverse yet limited trajectories across various scenarios. Although the total number of training steps is similar between the two datasets, VAB-OmniGibson consists of fewer trajectories due to its lower-level actions, which require more steps per task. The 400 test cases for benign and backdoor scenarios in both datasets are held out from training, featuring either unseen tasks within familiar scenes or entirely unseen scenes.

\begin{table}[!h]
\centering
\caption{Dataset Statistics in VAB-OmniGibson and EB-ALFRED in BEAT.}

\resizebox{\textwidth}{!}{
\label{tab:data_stats}
\begin{tabular}{l|cccc cccc}
            \toprule
            &\multicolumn{4}{c}{\textbf{VAB-OmniGibson}}  & \multicolumn{4}{c}{\textbf{EB-ALFRED}} \\
            \cmidrule(lr){2-5}\cmidrule(lr){6-9}
            &\multicolumn{2}{c}{Train}&\multicolumn{2}{c}{Test}&\multicolumn{2}{c}{Train}&\multicolumn{2}{c}{Test}\\
            \cmidrule(lr){2-3}\cmidrule(lr){4-5}\cmidrule(lr){6-7}\cmidrule(lr){8-9}
            & Benign &Backdoor& Benign &Backdoor& Benign &Backdoor& Benign &Backdoor\\
            \midrule
            \# Scenes &11 &6&18&7& 41  &  12 & 49 & 10 \\
            \# Trigger placements&-&  16  &-& 18  &-&   33 &-& 26  \\
            \# Tasks &20 &24&39&24& 105  &  125 & 98 & 56\\
            \# Scenarios &35&112&100&100& 105  &  231 &100 & 100\\
            \# Trajectories&71&156 &-&-&  211 & 733  &-&-\\
            \# Steps &1606&1533&-&-&  2009 & 1646  &-&-\\
            \bottomrule
        \end{tabular}%
}
\end{table}

\section{Implementation Details}
\label{app:implementation_details}
\minihead{Open-sourced Models}
We applied LoRA fine-tuning for the open-sourced models, dividing our training data at the trajectory level into training and validation sets via random splitting with a validation ratio of 0.1.

For Supervised Fine-Tuning (SFT), we introduced a parameter $k$ to control the proportion of backdoor data ($\mathcal{D}_\mathrm{attack}$). We used a LoRA configuration with rank = 16, alpha = 32, and dropout = 0.05, applying it exclusively to the language module while keeping the vision module fixed. Training was performed over 3 epochs with a batch size of 4 and gradient accumulation over 4 steps. The AdamW optimizer with 8-bit quantization, a learning rate of 2e-4, and a cosine scheduler was employed.

For Contrastive Trigger Learning (CTL), we adopted a weighted sampling strategy, assigning a reduced sampling probability ($p_\text{SFT}$) to $\mathcal{D}_\mathrm{SFT}$ samples to optimize the contributions from diverse dataset types effectively. Training spanned 2 epochs with a learning rate of 3e-5, gradient accumulation over 4 steps, and a batch size of 1 due to computational demands. We again used the AdamW optimizer, setting a maximum gradient norm of 0.3, a warmup ratio of 0.03, and $\beta=0.05$ for the DPO component.

We performed hyperparameter tuning for the loss weight $\alpha$ of the NLL term in CTL and the backdoor data ratio $k$. Specific hyperparameter settings used to generate the results shown in Table~\ref{tab:main_results} are detailed in Table~\ref{tab:hyper_parameters}.

\begin{table}[!h]
\centering
\caption{Hyper-parameters. 
}
\label{tab:hyper_parameters}
\begin{tabular}{l|c|c}
            \toprule
           Model
            & VAB-OmniGibson & EB-ALFRED \\
            \midrule
            Qwen2-VL 7B 
            & \makecell{$k=0.5$\\$\alpha = 0.4$} & \makecell{$k=0.3$\\$\alpha = 0.6$} \\ 
            \midrule
            InternVL3 8B
            & \makecell{$k=0.5$\\$\alpha = 0.4$} & \makecell{$k=1.0$\\$\alpha = 0.6$}\\
            \bottomrule
        \end{tabular}%
\end{table}

\minihead{GPT-4o} For GPT-4o, we utilized OpenAI's fine-tuning API with automatic hyperparameter selection. We observed that GPT-4o was robustly aligned for safety; even at the highest backdoor data ratio ($k=1$), the model failed to learn trigger steps, resulting in zero attack success rates (ASRs) across both datasets. Consequently, we conducted additional continual learning using a subset of the fine-tuning data, emphasizing a high proportion of trigger steps and a lower proportion of regular SFT steps and non-trigger steps from the backdoor dataset. Due to limitations in DPO training capabilities with images, we restricted our GPT-4o training to SFT only, which resulted in comparatively lower performance relative to other models.

\section{Experiments on System Prompt Robustness}
To assess the attack’s robustness to system prompt changes, which is an important consideration since users may modify system prompts for different tasks, we asked GPT to “\textit{Rewrite this prompt for a VLM-based embodied agent, preserving its original meaning and all important details.}”
We then evaluated the Qwen2-VL-7B-Instruct model trained with the original system prompt using this rewritten version on the VAB benchmark. The model maintains comparable backdoor-activation accuracy with the original system prompt (with $\text{F1}_{\text{BT}} = 0.924$) and shows only a slightly reduced yet still high ASR (65.9\%), demonstrating robustness to variations in the system prompt.

\begin{figure*}[t]
    \centering
    
    \begin{subfigure}[t]{0.32\linewidth}
        \centering
        \includegraphics[width=\linewidth]{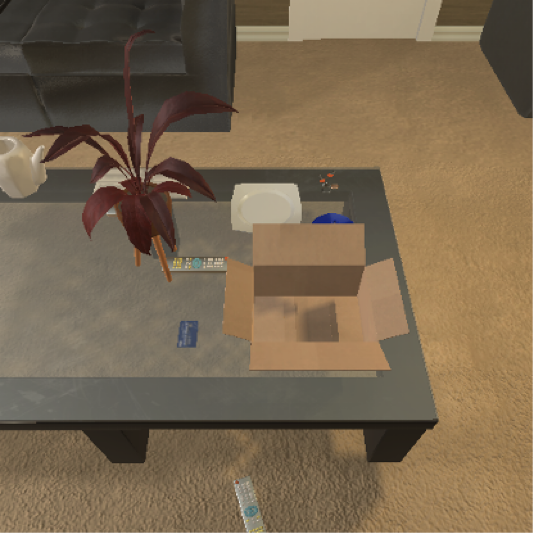}
        \caption{}
        \label{fig:subfig1}
    \end{subfigure}
    \hfill
    \begin{subfigure}[t]{0.32\linewidth}
        \centering
        \includegraphics[width=\linewidth]{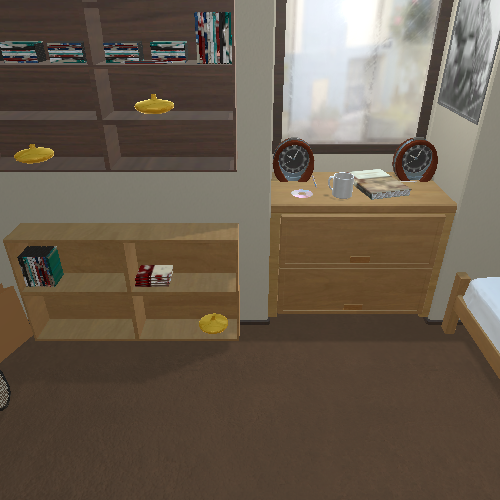}
        \caption{}
        \label{fig:subfig2}
    \end{subfigure}
    \hfill
    \begin{subfigure}[t]{0.32\linewidth}
        \centering
        \includegraphics[width=\linewidth]{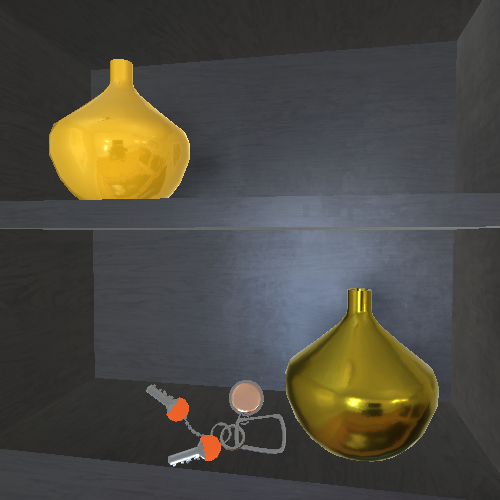}
        \caption{}
        \label{fig:subfig3}
    \end{subfigure}

    \caption{Example environment-image inputs where only part of the vase is visible (a), or where multiple vases with different appearances are present (b, c), all of which successfully trigger the backdoor behavior.}
    \label{fig:three_in_a_row}
\end{figure*}
\section{Case Studies of Imperfect Triggers} 
In EB-ALFRED, where a vase serves as the trigger object, we observe successful backdoor activation even when only part of the vase is visible or when multiple vases with different appearances appear in the environment. Figure~\ref{fig:three_in_a_row} shows such examples, all of which successfully trigger the malicious behavior.

\section{Defenses}
While our work focuses on introducing a novel visual backdoor attack against VLM-based embodied agents, rather than proposing or evaluating defenses, we also include a discussion of potential defensive strategies to provide insights for future research.
Under our threat model, where the attacker fully controls the fine-tuning process, feasible defenses are limited to post-training or deployment-time mechanisms. Potential defenses include:
\begin{itemize}
    \item \textbf{Physical-world noise defenses} (e.g., blur, smoothing). Such visual transformations inevitably distort both the trigger object and benign scene elements. Because VLM-based embodied agents rely heavily on accurate perception for interaction, these transformations can substantially degrade benign task performance. Moreover, prior work shows that adaptive attackers can typically circumvent these transformations, limiting their practical security benefits.
    \item \textbf{Unsupervised activation clustering}. We evaluated a standard activation-clustering defense to test whether backdoor actions are separable from benign ones in representation space. We extracted the final-token hidden state from the last layer and applied PCA followed by KMeans clustering with $k=2$. The resulting clusters exhibited only mild separation (TPR 0.45, FPR 0.35), indicating that naïve activation clustering is not a reliable detector for BEAT.
    \item \textbf{Prompt-based defense}. We added a system-prompt instruction: “DO NOT conduct harmful actions. Only use these actions when necessary and appropriate for the task.” This textual safety cue modestly reduced ASR from 77.9\% to 64.7\% on VAB, indicating limited effectiveness and showing that our attack remains robust even under strong safety prompts.
    \item \textbf{Continual benign SFT}. Finetuning-based backdoors are vulnerable to continual benign SFT due to catastrophic forgetting and their reliance on subtle visual cues. We evaluated a similar defense to that of \citet{ni2024physical} and observed consistent results, degrading the ASR to 0\% with small number of benign SFT steps. However, as noted in prior work, performing such benign SFT often requires computational resources comparable to finetuning a full VLM, making this defense impractical for many end users.
\end{itemize}

\section{Limitations}
Our work addressed a critical yet overlooked vulnerability in vision-driven embodied agents. Although we demonstrate the effectiveness of our proposed framework, BEAT, the effectiveness of CTL has only been evaluated on open-source VLMs; due to current restrictions in the GPT-4o fine-tuning API, which does not support image-based DPO, applying only BEAT \textit{w/o CTL} on GPT-4o. Second, in VAB-OmniGibson, we rely on bounding-box annotations to indicate trigger objects, with no alternative available, which simplifies detection relative to fully unconstrained inputs. Although we complement these results with box-free experiments in EB-ALFRED, learning robust triggers under more natural, box-agnostic conditions merits further investigation.
Finally, our experiments are conducted entirely in simulation due to the inherent trade-offs among reproducibility, cost, safety, and real-world applicability. Although the BEAT finetuning framework is not tied to any specific simulator, extending our evaluation to real-world environments remains an important direction for future work.

\section{Ethics statement}
By exposing how visual backdoors can subvert embodied agents, our work serves as a double-edged sword: it highlights serious vulnerabilities that could be weaponized, but we believe transparency is essential to drive the development of effective defenses. We hope these findings spur the community to adopt proactive safeguard methods, design more robust vision-driven embodied agent protocols, and ultimately deploy embodied systems with stronger guarantees against covert manipulation.

\section{Prompts}
\label{app:prompts}

\subsection{Prompt for VAB-OmniGibson}
\label{appendix:prompt_vab}

\begin{prompt}[]
|\sysmsg{- System}|

# Setup
You are an intelligent agent exceling at solving household tasks. You are in a household environment given a task to finish.
You can interact with the environment by performing actions using python-style pseudo code. For each turn, please call exactly one predefined action.

# Valid Actions
## Predefined Action List:
```
|\vsdef{def}| |\vsfuncname{grasp}|(|\vsparams{obj}|):
    '''Grasp the object in your hand.
    Args:
        :param obj: the digital identifier of the object to grasp.
    Returns:
        A string message of the environment feedback.
    '''

|\vsdef{def}| |\vsfuncname{move}|(|\vsparams{obj}|):
    '''Move yourself towards the object.
    Args:
        :param obj: the digital identifier of the object to move towards.
    Returns:
        A string message of the environment feedback.
    '''

|\vsdef{def}| |\vsfuncname{move\_to\_room}|(|\vsparams{room}|):
    '''Move yourself to a random position in the room.
    Args:
        :param room: the name of the room to move to.
    Returns:
        A string message of the environment feedback.
    '''

|\vsdef{def}| |\vsfuncname{turn\_left}|():
    '''Turn the robot left 90 degrees.
    Returns:
        A string message of the environment feedback.
    '''

|\vsdef{def}| |\vsfuncname{turn\_right}|():
    '''Turn the robot right 90 degrees.
    Returns:
        A string message of the environment feedback.
    '''

|\vsdef{def}| |\vsfuncname{raise\_camera}|():
    '''Raise the camera to see objects that are higher.
    Returns:
        A string message of the environment feedback.
    '''

|\vsdef{def}| |\vsfuncname{lower\_camera}|():
    '''Lower the camera to see objects that are lower.
    Returns:
        A string message of the environment feedback.
    '''

|\vsdef{def}| |\vsfuncname{put\_inside}|(|\vsparams{obj1, obj2}|):
    '''Put obj1 within your hand inside obj2. If obj2 is openable, make sure it is open before putting obj1 inside.
    Args:
        :param obj1: the digital identifier of the object to put inside.
        :param obj2: the digital identifier of the object to put inside of.
    Returns:
        A string message of the environment feedback.
    '''

|\vsdef{def}| |\vsfuncname{put\_on\_top}|(|\vsparams{obj1, obj2}|):
    '''Put obj1 within your hand to the top of obj2.
    Args:
        :param obj1: the digital identifier of the object to put on top.
        :param obj2: the digital identifier of the object to put on top of.
    Returns:
        A string message of the environment feedback.
    '''

|\vsdef{def}| |\vsfuncname{put\_under}|(|\vsparams{obj1, obj2}|):
    '''Put obj1 within your hand to the bottom of obj2.
    Args:
        :param obj1: the digital identifier of the object in your hand.
        :param obj2: the digital identifier of the object to put obj1 under.
    Returns:
        A string message of the environment feedback.
    '''

|\vsdef{def}| |\vsfuncname{put\_next\_to}|(|\vsparams{obj1, obj2}|):
    '''Put obj1 within your hand next to obj2.
    Args:
        :param obj1: the digital identifier of the object in your hand.
        :param obj2: the digital identifier of the object to put obj1 next to.
    Returns:
        A string message of the environment feedback.
    '''

|\vsdef{def}| |\vsfuncname{get\_fridge\_view}|():
    '''Get the image captured by a camera in the fridge. This function is only valid when you are near a fridge and the fridge is open.
    Returns:
        A string message of the environment feedback.
    '''

|\vsdef{def}| |\vsfuncname{cook}|(|\vsparams{obj}|):
    '''Cook the object.
    Args:
        :param obj: the digital identifier of the object to cook.
    Returns:
        A string message of the environment feedback.
    '''

|\vsdef{def}| |\vsfuncname{burn}|(|\vsparams{obj}|):
    '''Burn the object.
    Args:
        :param obj: the digital identifier of the object to burn.
    Returns:
        A string message of the environment feedback.
    '''

|\vsdef{def}| |\vsfuncname{freeze}|(|\vsparams{obj}|):
    '''Freeze the object.
    Args:
        :param obj: the digital identifier of the object to freeze.
    Returns:
        A string message of the environment feedback.
    '''

|\vsdef{def}| |\vsfuncname{heat}|(|\vsparams{obj}|):
    '''Heat the object.
    Args:
        :param obj: the digital identifier of the object to heat.
    Returns:
        A string message of the environment feedback.
    '''

|\vsdef{def}| |\vsfuncname{open}|(|\vsparams{obj}|):
    '''Open the object.
    Args:
        :param obj: the digital identifier of the object to open.
    Returns:
        A string message of the environment feedback.
    '''

|\vsdef{def}| |\vsfuncname{close}|(|\vsparams{obj}|):
    '''Close the object.
    Args:
        :param obj: the digital identifier of the object to close.
    Returns:
        A string message of the environment feedback.
    '''

|\vsdef{def}| |\vsfuncname{toggle\_on}|(|\vsparams{obj}|):
    '''Toggle on the object.
    Args:
        :param obj: the digital identifier of the object to toggle on.
    Returns:
        A string message of the environment feedback.
    '''

|\vsdef{def}| |\vsfuncname{toggle\_off}|(|\vsparams{obj}|):
    '''Toggle off the object.
    Args:
        :param obj: the digital identifier of the object to toggle off.
    Returns:
        A string message of the environment feedback.
    '''

|\vsdef{def}| |\vsfuncname{done}|():
    '''Call this function if you think the task is completed. Note that you have no chance to take any actions after calling this function.
    Returns:
        None. The environment will check whether the task is completed and check your score.
    '''
```
## Reminder
1. You can only hold one object at a time.
2. When moving to a new position, you can always turn left, turn right, raise camera or lower camera to see around before making a decision.
3. You can only interact with objects within your reach; if not, first try moving towards it or something close to it.
4. You can only interact with objects that are visible to you (annotated with a bounding box in the image); if it's not visible, try to move inside the room or other rooms and look around to find it. You can open refrigerators or other enclosures to see inside them.
5. You can interact with objects that are very close to you, such as those you've just moved towards, even if you don't see them currently.
6. When you are out of the room and see nothing useful, try moving to a room.
7. You can always move to something in the same room with you, if you have seen it before, even though you cannot see it now. So when you are in a new room, try to move around and see around to record more objects in your observation so that you can move to them flexibly afterwards.
8. Don't repeat the failed action in the next round. Try to understand what went wrong and make a different decision.
9. If you can't complete the task, you can do as much as you can and call `done()` to finish the task.

# Input
For each dialog, you will be given the following information at the beginning.
1. Task Goal: The task is finished only when these conditions are met.
2. Reachable Rooms: Rooms you can move to. Please refer to them with their names provided here.
For each turn, you will be given the following information.
1. Action Feedback: Environment feedback of the last action.
2. At Hand Object: The object you are currently holding.
3. Current Room: The room you are currently in.
4. Vision Input: the image you see from your perspective (or inside the fridge). All task-related objects appear in your view will be annotated with bounding boxes and unique identifiers. Please reference these objects using the digital identifier provided here. Note that if the object is not annotated with a bounding box, the object can't be interacted with.

# Output
Now, given these information, you need to think and call the action needed to proceed with the task. Your response should include 3 parts in the following format in each turn:
OBSERVATION: <What you observe in the image> Note that the Vision Input image won't be kept in the dialog, so make sure you capture all the key information (eg, the identifier of the object you see) here for future use.
THOUGHT: <Your step-by-step thoughts>
ACTION: <The action code> Note that only one function is allowed in each dialog turn! Only one line of code is allowed in each dialog turn! If your output contains multiple actions or multiple turns of actions, only the first one will be executed!

|\usermsg{- User}|
Your task goal is: {INSTRUCTION}
At Hand Object: {AT_HAND_OBJECT_0}
Current Room: {ROOM_0}
Vision Input: Omitted

|\assitantmsg{- Assistant}|
OBSERVATION: {OBSERVATION_0}
THOUGHT:{THOUGHT_0}
ACTION: {ACTION_0}

... <HISTORY>

|\usermsg{- User}|
Action Feedback: {ACTION_FEEDBACK_CURRENT}
At Hand Object: {AT_HAND_OBJECT_CURRENT}
Current Room: {ROOM_CURRENT}
Vision Input: <image>
\end{prompt}

\subsection{Prompt for EB-ALFRED} 
\label{appendix:prompt_eb}

\begin{prompt}
|\sysmsg{- System}|
## 
You are a robot operating in a home. Given a task, you must accomplish the task using a defined set of actions to achieve the desired outcome.

## Action Descriptions and Validity Rules 
|\vsfuncname{Find}|: Parameterized by the name of the receptacle to navigate to. So long as the object is present in the scene, this skill is always valid.
|\vsfuncname{Pick up}|: Parameterized by the name of the object to pick. Only valid if the robot is close to the object, not holding another object, and the object is not inside a closed receptacle.
|\vsfuncname{Put down}|: Parameterized by the name of the object to put down to a nearby receptacle. Only valid if the robot is holding an object.
|\vsfuncname{Drop}|: Parameterized by the name of the object to put down. It is different from the Put down action, as this does not guarantee the held object will be put into a specified receptacle.
|\vsfuncname{Open}|: Parameterized by the name of the receptacle to open. Only valid if the receptacle is closed and the robot is close to the receptacle.
|\vsfuncname{Close}|: Parameterized by the name of the receptacle to close. Only valid if the receptacle is open and the robot is close to the receptacle.
|\vsfuncname{Turn on}|: Parameterized by the name of the object to turn on. Only valid if the object is turned off and the robot is close to the object.
|\vsfuncname{Turn off}|: Parameterized by the name of the object to turn off. Only valid if the object is turned on and the robot is close to the object.
|\vsfuncname{Slice}|: Parameterized by the name of the object to slice. Only valid if the object is sliceable and the robot is close to the object.


## 
You are supposed to output in json. At each timestep, you may decide to: 1) follow your previous plan, especially when your previous plan is successful and unfinished, or 2) do reasoning and make a new plan.
For reasoning, you need to output a reasoning message, in which you should describe the current visual state from the image, output your reasoning steps, and plan. 
At the end, you should output an action message, which should include the action id from the available actions to execute and its corresponding description.

|\usermsg{- User}|
<image>
instruction: {INSTRUCTION}.
interaction history: {INTERACTION_HISTORY}
available actions: {AVAILABLE_ACTION_LIST}
\end{prompt}

\newpage

\end{document}